\documentclass[lettersize,journal]{IEEEtran}
\usepackage{cite}
\usepackage{amsmath,amssymb,amsfonts}
\usepackage{algorithmic}
\usepackage{relsize}
\usepackage{graphicx}
\usepackage{textcomp}
\usepackage{xcolor}
\usepackage{algorithm,algorithmic}

\usepackage{subcaption} 
\usepackage{soul}
\usepackage{caption}
\usepackage{hyperref}
\usepackage{cancel}
\usepackage{soul}
\usepackage{array}
\usepackage{booktabs} 
\usepackage{amssymb} 
\usepackage{pifont} 
\usepackage{graphicx} 

\newcommand{\cmark}{\ding{51}}
\newcommand{\xmark}{\ding{55}}
\ifCLASSINFOpdf
\else
\fi

\begin{document}

\title{Multi-Agent DRL for Queue-Aware Task Offloading in Hierarchical MEC-Enabled Air-Ground Networks}

\author{Muhammet Hevesli,~\IEEEmembership{Member,~IEEE,} Abegaz Mohammed Seid,~\IEEEmembership{Member,~IEEE,}
        Aiman Erbad,~\IEEEmembership{Senior Member,~IEEE,}  Mohamed Abdallah,~\IEEEmembership{Senior Member,~IEEE}%

\thanks{Muhammet Hevesli, Abegaz Mohammed Seid  and Mohamed Abdallah are with the Division of Information and Computing Technology,
College of Science and Engineering, Hamad Bin Khalifa University, Doha, Qatar (\textit{corresponding authors: Muhammet Hevesli, and Abegaz Mohammed Seid)}, e-mail: muhe45212@hbku.edu.qa; mabegaz@hbku.edu.qa; moabdallah@hbku.edu.qa).} 
\thanks{Aiman Erbad is with the College of Engineering, Qatar University, Doha, Qatar (e-mail: aerbad@qu.edu.qa).} 
\thanks{This work was made possible by NPRP-Standard (NPRP-S) Thirteen (13th) Cycle grant NPRP13S-0205-200265 from the Qatar National Research Fund. The findings achieved herein are solely the responsibility of the authors.}
}

\markboth{IEE TCCN 2024}%
{Shell \MakeLowercase{\textit{\textit{et al}.}}: Bare Demo of IEEEtran.cls for IEEE Journals}

\maketitle

\begin{abstract} Mobile edge computing (MEC)-enabled air-ground networks advance 6G wireless networks by utilizing aerial base stations (ABSs) such as unmanned aerial vehicles (UAVs) and high altitude platform stations (HAPS) to provide dynamic services to ground IoT devices (IoTDs). These IoTDs support real-time applications like multimedia and Metaverse services, which demand high computational resources and strict quality of service (QoS) guarantees, specifically in terms of latency and efficient task queue management. However, IoTDs often face constraints in energy and computational power, requiring efficient queue management and task scheduling to maintain QoS. To address these challenges, UAVs and HAPS are deployed to supplement the computational limitations of IoTDs by offloading tasks for distributed processing. Due to UAVs' resource limitations, particularly in terms of power and coverage area, HAPS are used to enhance their capabilities and extend coverage. Overloaded UAVs may relay tasks to HAPS, creating a multi-tier computing system. This paper addresses the overall energy minimization problem in the MEC-enabled air-ground integrated network (MAGIN) by optimizing UAV trajectories, computing resource allocation, and queue-aware task offloading decisions. The optimization problem is highly complex due to the nonconvex and nonlinear nature of this hierarchical system, which traditional methods cannot effectively solve. To tackle this, we reformulate the problem as a multi-agent Markov decision process (MDP) with continuous action spaces and heterogeneous agents. We propose a novel variant of multi-agent proximal policy optimization (MAPPO) with Beta distribution (MAPPO-BD) to solve this problem. Extensive simulations show that MAPPO-BD significantly outperforms other baselines, achieving superior energy savings and efficient resource management in MAGIN, while adhering to constraints related to queue delays and edge computing capabilities. \end{abstract}
\begin{IEEEkeywords}
Mobile edge computing, multi-agent deep reinforcement learning, edge network, air-ground network, Metaverse.
\end{IEEEkeywords}
\IEEEpeerreviewmaketitle
\section{Introduction}
\IEEEPARstart{T}he advent of sixth-generation (6G) wireless networks, driven by the rapid advancement of Internet of Things (IoT) technology and the extensive deployment of 5G networks, is shaping the future of telecommunications \cite{8782879},\cite{9144301}. These emerging networks aim to provide multidimensional, intelligent, and green communication features, fostering ubiquitous connectivity among a vast array of devices \cite{zhang20196g}. However, the increasing computational demands of real-time applications like autonomous driving, Metaverse services \cite{jiang2023qoe},\cite{articlemeta}, and telemedicine, coupled with the limited computational capabilities and battery capacities of local devices, present significant challenges. Mobile edge computing (MEC), as an extension of cloud computing, emerges as a key solution \cite{mao2017survey}. MEC brings computing resources closer to the network edge, reducing data travel distance and enabling faster processing with lower latency. By offloading tasks to edge servers, MEC decreases the energy consumption of IoT devices (IoTDs), extending their battery life\cite{siriwardhana2021survey}.

In the evolving landscape of 6G and IoT, terrestrial edge servers often face challenges such as blockage and limited coverage for remote users \cite{Liu2019},\cite{Han2020}. Unmanned aerial vehicles (UAVs) have emerged as a viable solution, offering high mobility, flexible deployment, and efficient computational offloading, thus enhancing MEC networks \cite{du2018energy}. They provide reliable communication links, particularly in emergency scenarios where terrestrial infrastructure is impractical \cite{du2019joint}. Furthermore, the integration of high altitude platform stations (HAPS) with UAVs in air-ground integrated networks (AGIN) offers an effective blend of wide coverage and enhanced communication, presenting a cost-effective alternative to satellite systems \cite{kurt2021vision}. This synergy between UAVs, HAPS, and ground components in 6G technology marks a significant shift towards more dynamic, spatially distributed computing and communication services \cite{jia2022hierarchical}.

In the context of MEC-enabled AGIN (MAGIN), the variability in task types and network conditions necessitates adaptive queuing mechanisms \cite{chen2021efficient}. These mechanisms must effectively prioritize tasks, manage network resources \cite{nagarajan2021effective}, and optimize processing times to maintain the network's integrity and performance. The integration of a multi-tier computing architecture \cite{el2019joint} comprising UAVs and HAPS introduces a groundbreaking approach to queue-aware hierarchical task offloading \cite{8594571}. This framework effectively synergizes the agility of UAVs with the extensive coverage provided by HAPS, establishing a multi-layered network topology that is essential for enhancing network efficiency and service delivery. UAVs, with their rapid deployment capabilities, cater to immediate and localized network demands, ensuring optimized coverage and minimized latency for time-sensitive tasks. In contrast, HAPS, operating at higher altitudes, create a stable and extensive network layer, crucial for offloading computationally intensive tasks from terrestrial servers and lower-tier UAVs. This hierarchical arrangement allows for intelligent task routing based on computational intensity and latency requirements \cite{qiu2019air}, aligning tasks with the most appropriate network tier. This is pivotal in meeting the dynamic and evolving demands of queue-aware task scheduling and offloading in environments with stochastic arrival of delay-sensitive and computationally intensive tasks.

Artificial intelligence (AI), mainly reinforcement learning (RL), has become a powerful tool to address complex optimization problems in 5G and  B5G networks, such as resource allocation, energy efficiency, and task offloading optimization. In MAGIN, RL provides effective solutions for resource allocation, especially in dynamic and unpredictable environments \cite{liao2021learning}, \cite{zhou2020deep}. The evolution of RL into deep reinforcement learning (DRL) signifies a substantial progression in this field, with DRL effectively handling continuous action spaces and addressing the limitations of traditional RL methods' discretization \cite{8714026}. However, DRL faces challenges in centralized execution within dynamic, large-scale networks, leading to inefficiencies in decision-making. To overcome this, the multi-agent deep reinforcement learning (MADRL) framework integrates centralized training with decentralized execution \cite{9485089}, catering to environments where IoTDs are constrained by energy and computing resources. The centralized training aspect of MADRL allows for a comprehensive understanding and learning from the collective experiences of multiple agents \cite{seid2021multi}, leading to more informed and sophisticated policy development. Concurrently,  decentralized execution ensures that each agent can independently make decisions in real-time, a critical factor in dynamic environments where swift and autonomous responses are essential \cite{9209079}. 

Despite significant advances in UAV-enabled MEC networks, current research reveals several limitations that hinder the full potential of hierarchical air-ground network architectures in 6G environments. Existing hierarchical frameworks \cite{jia2022hierarchical, ren2022caching} often neglect dynamic queue management and efficient UAV-HAPS coordination, leading to suboptimal resource allocation and increased energy consumption. Additionally, the absence of adaptive, queue-aware task offloading and trajectory planning exacerbates latency issues in real-time applications. While some studies \cite{seid2021multi, qin2022optimal} utilize MADRL, they often treat UAVs as homogeneous agents, overlooking their varied roles and interactions with HAPS. Motivated by the aforementioned research scopes and to address these limitations, our work integrates a MAGIN with an adaptive, queue-aware task offloading strategy and a customized heterogeneous MADRL approach. This integration aims to enhance the QoS of IoTDs, minimize energy consumption, and ensure effective task management for real-time applications in 6G wireless networks. By focusing on total energy minimization, our approach ensures the efficient and sustainable operation of the network, allowing it to adapt to varying network conditions and heterogeneous IoTD demands while maintaining real-time performance. In this work, we consider task latency and queue management as the primary QoS parameters. Latency ensures timely task completion, while efficient queue management prevents task delays and excessive backlogs. We deploy \textbf{multi-tier computing approach within MAGIN}, which integrates multiple layers of computational resources, including UAVs and HAPS, to create a hierarchical structure. This system enhances the network's capabilities and coverage, particularly by overcoming the power and computing resources limitations of UAVs. Alongside this, we introduce a \textbf{flexible queue-aware task offloading strategy} that adjusts to changing task demands and the different capacities of the network. This method improves how tasks are scheduled, and resources are allocated, making the network more efficient. Additionally, we develop a \textbf{tailored heterogeneous multi-agent DRL (HMADRL) method}, using the multi-agent proximal policy optimization with Beta distribution (MAPPO-BD) algorithm. This strategy is designed to effectively optimize task offloading decisions, UAV flight paths, and computing resource distribution to minimize MAGIN's total energy consumption. The main contributions of our work are as follows: \begin{itemize}
\item Multi-tier Computing in MAGIN: Our study presents an innovative approach to enhance air network capabilities and coverage by integrating HAPS with UAVs. This integration enables IoTDs to connect directly with HAPS when UAVs are unavailable or face computational resources and power limitations.

\item Adaptive Queue-Aware Task Offloading in a Multi-tier MAGIN: In this work, we address the dynamic and complex nature of task arrivals and the varying capacities of IoTDs, UAVs, and HAPS by developing an adaptive queue-aware multi-tier task offloading strategy. This mechanism intelligently adjusts task offloading decisions and computing resource allocations in real time, informed by current queue statuses. 

\item Energy-Efficient Task Offloading in Multi-tier MAGIN: We formulate the Joint multi-UAV Trajectory, Queue-aware task Offloading, and Resources Allocation (JUTQORA) optimization problem to minimize the overall energy consumption in the multi-tier MAGIN, adhering to the constraints of queue delays and edge servers computing capabilities. The optimization problem is a mixed integer nonlinear programming (MINLP) challenge, adding to its non-convexity.
    
\item Customized HMADRL Framework: To tackle the formulated optimization problem, we transform it into a multi-agent MDP leveraging three types of heterogeneous agents: IoTD agents, UAV agents, and HAPS agents. Then, we propose a novel MAPPO-BD-based algorithm, utilizing the Beta distribution in the actor networks instead of the conventional Gaussian distribution. This approach enhances the performance of heterogeneous agents with varying action boundaries, facilitating uniform exploration and avoiding boundary effects.

\item Evaluation analysis: Through extensive simulations, we demonstrate the superiority of our MAPPO-BD algorithm by comparing it against baselines such as multi-agent deep deterministic policy gradient (MADDPG), MAPPO utilizing normal distribution (MAPPO-ND), and partial optimization with MAPPO framework (PO-MAPPO-BD). The results show that MAPPO-BD significantly outperforms these baselines, achieving superior energy savings and more efficient resource utilization in MAGIN.
\end{itemize}

The rest of this paper is organized as follows. Section II discusses the related work. Section III presents the system model and the optimization problem. We provide a solution to the problem in Section IV. Sections V and VI, respectively, provide detailed descriptions of the numerical findings and conclusions.

\section{Related Works}
Recently, in the context of UAV-enabled MEC networks, research has primarily focused on enhancing MEC service provisioning to remote IoTDs. The deployment of multi-UAVs in MEC networks represents a significant advancement in network optimization. Huang \textit{et al}. \cite{huang2021multi} delved into the optimization of computation offloading, channel allocation, power control, and computation resource allocation in small cell networks using MADRL. This study demonstrated the critical role of UAVs in augmenting MEC capabilities and improving network efficiency. Shi \textit{et al}. \cite{SHI2024103371} explored collaborative UAV-assisted MEC, where multiple UAVs worked in tandem to offload computational tasks, utilizing a MADRL algorithm for decision-making. Similarly, Seid \textit{et al}. \cite{seid2021multi} proposed a clustered multi-UAV network to minimize computation costs and maintain QoS in IoT networks, again employing MADRL for optimization. These studies highlight the complexities and opportunities in managing multi-UAV MEC networks.

The integration of HAPS with multiple UAVs in MEC networks offers a new dimension in network enhancement. Ren \textit{et al}. \cite{ren2022caching} proposed a three-layer computation framework that utilized HAPS, terrestrial network edges, and UAVs. This work aimed to minimize communication delays through strategic caching, thereby improving the efficiency of task execution in MEC networks. Adding to this, Qin \textit{et al}. \cite{qin2022optimal} proposed a power IoT (PIoT) heterogeneous system model that integrated HAPS, UAVs, and ground IoTDs. This model emphasized the efficient use of subchannels, task splitting, and computational resource allocation in heterogeneous networks. By utilizing non-orthogonal multiple access protocols, the model enabled subchannel reuse among multiple UAVs, thereby enhancing energy efficiency and optimizing queue-aware resource allocation throughout the network. Lakew \textit{et al}. \cite{lakew2022intelligent} explored a dynamic, heterogeneous aerial IoT network that integrates HAPS, multiple UAVs, and IoTDs in areas with limited service. They focused on optimizing IoTD association, partial offloading, and resource allocation to enhance service quality and reduce energy usage. Their methodology utilized a MADDPG-based algorithm, combining centralized training with decentralized execution. These studies collectively demonstrate the evolving complexity and potential of HAPS-UAV synergies in MEC-enabled networks.

Hierarchical and queue-aware task offloading strategies are pivotal for efficient network management in UAV-enabled MEC systems. UAVs have the capability to compute part of the tasks while offloading the remaining tasks to other edge nodes in the network. Huang \textit{et al}. \cite{huang2021task} tackled this issue by proposing a collaborative approach between UAVs and ground fog nodes. Their method involves a two-stage offloading strategy that optimizes UAV trajectory, transmission power, and computation offloading ratios, all while maintaining QoS requirements. A novel approach by Yu \textit{et al}. \cite{yu2020joint} presented a UAV-enabled edge network to aid IoTDs obstructed by terrestrial signal blockages in relaying part of their tasks to edge cloud nodes. This study addressed the optimization of IoT task offloading and UAV placement to minimize service delay and maximize UAV energy efficiency. Additionally, Jia \textit{et al}. \cite{jia2022hierarchical} introduced a hierarchical aerial computing framework comprising both HAPS and UAVs. In this framework, HAPS supports UAVs in computing-intensive tasks, enhancing MEC service for terrestrial IoTDs. Their MADRL-based algorithm aimed to maximize the total IoT data processed by optimizing device associations and offloading ratios. These studies underscore the significance of strategic task offloading in complex, layered network environments. Extending this concept further, Liao \textit{et al}. \cite{liao2021learning} focused on minimizing the long-term energy consumption of IoTDs in a space-air-ground-integrated network that integrates satellites with multi-UAVs. Their approach involved a learning-based algorithm for queue-aware task offloading and computational resource allocation, jointly optimizing task splitting and offloading.
\begin{table*}[ht]
\centering \fontsize{7}{7}\selectfont 
\caption{Summary of the relevant literature.}
\label{relatedworks}
\begin{tabular}{@{}l>{\centering\arraybackslash}m{0.8cm}>{\centering\arraybackslash}m{0.8cm}>{\centering\arraybackslash}m{1.4cm}>{\centering\arraybackslash}m{0.8cm}>{\centering\arraybackslash}m{1.3cm}>{\centering\arraybackslash}m{1.4cm}>{\centering\arraybackslash}m{1.6cm}>{\centering\arraybackslash}m{1.2cm}>{\centering\arraybackslash}m{1.4cm}>{\centering\arraybackslash}m{1.5cm}@{}}
\toprule
\textbf{Ref.} & \textbf{Single UAVs} & \textbf{Multi-UAVs} & \textbf{Multi-tier Computing} & \textbf{Queue Aware} & \textbf{Partial \quad Offloading} & \textbf{Trajectory Plan} & \textbf{Interference Management} & \textbf{Resource Allocation} & \textbf{Homo. MADRL} & \textbf{Hetero. MADRL}\\
\midrule
\cite{huang2021multi} & \xmark & \xmark & \xmark & \xmark & \cmark & \xmark & \cmark & \cmark & \cmark & \xmark  \smallskip \\ \hline
\cite{yu2020joint} & \cmark & \xmark & \cmark & \xmark & \cmark & \cmark & \cmark & \cmark & \xmark & \xmark \smallskip \\ \hline
\cite{huang2021task} & \cmark & \xmark & \cmark & 
\xmark & \cmark & \cmark & \cmark & \xmark & 
\xmark & \cmark \smallskip  \\ \hline
\cite{ren2022caching} & \xmark & \xmark & \xmark & \xmark & \cmark & \xmark & \cmark & \cmark & \xmark & \cmark \smallskip \\ \hline
\cite{seid2021multi} & \xmark & \cmark & \xmark & \xmark & \xmark & \xmark & \cmark & \cmark & \cmark & \xmark \smallskip  \\ \hline
\cite{jia2022hierarchical} & \xmark & \cmark & \cmark & \xmark & \xmark & \xmark & \cmark & \xmark & \xmark & \xmark \smallskip \\ \hline
\cite{SHI2024103371} & \xmark & \cmark & \xmark & \xmark & \xmark & \cmark & \xmark & \xmark & \cmark & \xmark \smallskip  \\ \hline
\cite{10314006} & \xmark & \cmark & \xmark & \xmark & \xmark & \cmark & \xmark & \xmark & \cmark & \xmark \smallskip \\ \hline
\cite{9209079} & \xmark & \cmark & \xmark & \xmark & \cmark & \cmark & \xmark & \xmark & \cmark & \xmark \smallskip \\ \hline
\cite{qin2022optimal} & \xmark & \cmark & \xmark & \cmark & \cmark & \xmark & \cmark & \cmark & \xmark & \xmark \smallskip \\ \hline
\cite{lakew2022intelligent} & \xmark & \cmark & \xmark & \xmark & \cmark & \xmark & \xmark & \cmark & \xmark & \cmark \smallskip \\ \hline
\cite{LI2024103341} & \xmark & \cmark & \xmark & \xmark & \xmark & \cmark & \xmark & \xmark & \xmark & \cmark \smallskip \\ \hline
\cite{liao2021learning} & \xmark & \cmark & \xmark & \cmark & \cmark & \xmark & \cmark & \cmark & \xmark & \xmark \smallskip \\ \hline
Our work & \xmark & \cmark & \cmark & \cmark & \cmark & \cmark & \cmark & \cmark & \xmark & \cmark \smallskip \\ \hline
\end{tabular}
\end{table*}
The application of homogeneous and heterogeneous MADRL in network optimization presents innovative approaches to managing complex scenarios. Wang \textit{et al}. \cite{9209079} used a MADRL-based algorithm in a homogeneous agent setting, focusing on optimizing UAV trajectories for network fairness and energy efficiency. Expanding on heterogeneous approaches, a study by Li \textit{et al}. \cite{LI2024103341} introduced a multi-UAV-assisted task offloading framework. This framework aimed to minimize UAV energy consumption and user task latency by optimizing UAV trajectories and task offloading strategies, employing a heterogeneous MADRL approach where UAVs and ground users are treated as separate types of agents. Additionally, in the context of smart agriculture, another study \cite{10314006} proposed a joint UAV task scheduling, trajectory planning, and resource-sharing framework for multi-UAV-assisted wireless sensor networks. This study aimed at minimizing energy consumption and network latency, leverages a MADRL-based algorithm to effectively manage UAV charging, data collection, and energy sharing with sensor nodes. 

As shown in Table \ref{relatedworks}, which outlines a comparative summary of our work and some other works from the literature, our work stands out by integrating multiple advanced features to enhance the performance of multi-tier computing in hierarchical MAGIN environments. Unlike existing studies, we leverage multi-UAVs with a flexible queue-aware task offloading strategy, advanced resource allocation, and effective trajectory planning. Our approach also pioneers the use of heterogeneous MADRL to solve the optimization problem in such a dynamic network.
\section{System Model}
In this study, we introduce a hybrid air-ground network model designed to enhance connectivity in challenging environments such as hotspots, disaster-stricken areas, or regions with temporary population surges where expanding terrestrial network infrastructure is impractical. The model incorporates a multi-tier architecture consisting of randomly distributed IoTDs, $M$ UAVs, and a single HAPS. The IoTDs, denoted by $\mathcal{N} = \{1, 2, \dots, N\}$, are uniformly distributed across the target area, ensuring consistent positioning within the network boundaries. The aerial base stations (ABSs) are represented by $\mathcal{M} = \{0, 1, 2, \dots, M\}$, with the index $m=0$ specifically referring to the HAPS. 

\begin{figure}[h]
    \centering
    \begin{subfigure}{0.49\textwidth}
        \centering
\includegraphics[width=\textwidth, height=5cm]{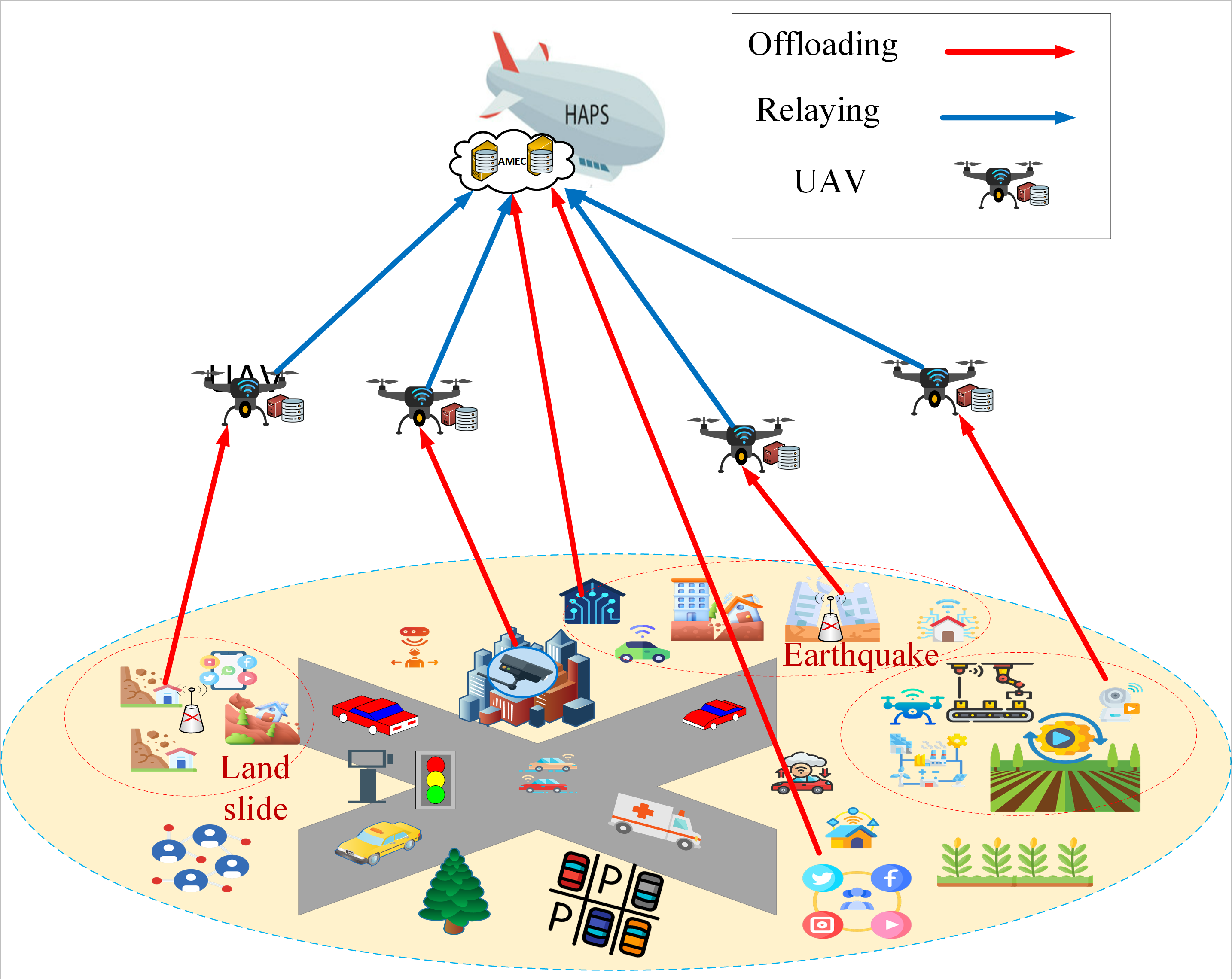}
        \caption{System model.}
        \label{fig:systemmodel-a}
    \end{subfigure}
    \hfill
    \begin{subfigure}{0.49\textwidth}
        \centering
\includegraphics[width=\textwidth, height=4.5cm]{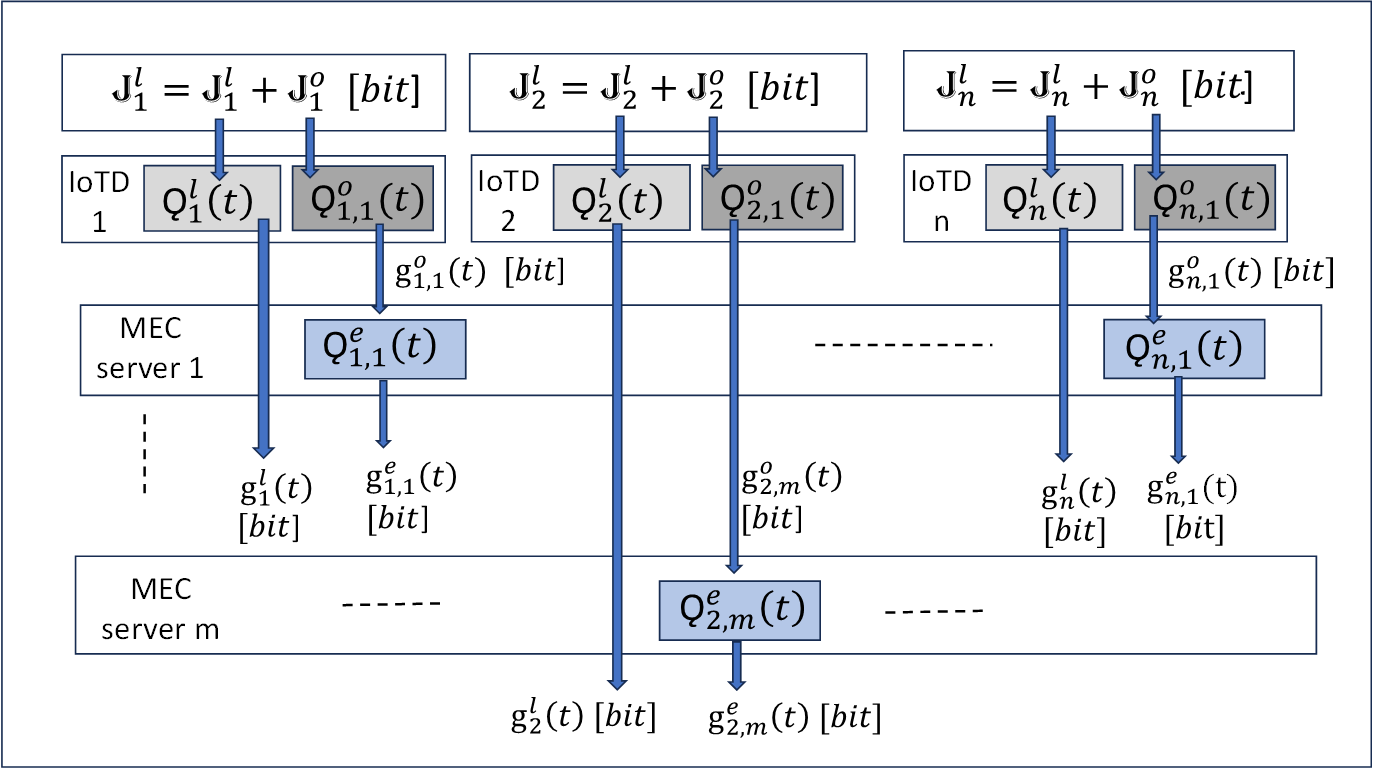}
        \caption{Queuing model.}
        \label{fig:systemmodel-b}
    \end{subfigure}
    \caption{System model and queuing model.}
    \label{fig:System Model}
\end{figure}

The HAPS is deployed at a fixed position above the operation area, serving as a stable, high-altitude relay that extends connectivity beyond the range of the UAVs. UAVs are positioned at a fixed altitude \( h_m \) and are crucial for dynamically adjusting network coverage to optimize service for distributed IoTDs. They operate either in flying or hovering mode. In flying mode, UAVs fly within the studied coverage area to respond to changing network demands and IoTD distributions and to bridge coverage gaps in evolving scenarios.
However, in hovering mode, UAVs maintain a fixed position to provide continuous coverage to areas with high data demand, especially for sustained operations. The UAVs adjust their positions and manage the offloading of computational tasks. If a UAV's resources are insufficient, tasks are further offloaded to the HAPS, ensuring efficient processing and data handling while maintaining high QoS. The multi-tier structure also allows IoTDs to connect directly to the HAPS when out of UAV range, enhancing the robustness and reach of the communication system. The proposed system model is illustrated in Fig. \ref{fig:System Model}. To enhance the readability, the main notations in this paper are summarized in Table \ref{tab:table1}.
\begin{table*}[ht]
\centering 
\caption{LIST OF KEY NOTATIONS.}\label{tab:table1}
\begin{tabular}{|c|l|c|l|}
\hline
\textbf{Symbol} & \textbf{Definition} & \textbf{Symbol} & \textbf{Definition}  \\ \hline
\( \theta_m(t) \) & Flight direction of UAV \( m \) & \( d_m(t) \) & Distance travelled by UAV \( m \) \\ \hline
\( X_m(t), Y_m(t) \) & Position coordinates of UAV \( m \) & \( x_m(t), y_m(t) \) & Position coordinates of IoTD \( n \) \\ \hline
\( d_{m,m'}(t) \) & Distance between UAV \( m \) and UAV \( m' \) & \( d_{\min} \) & Min. allowable distance between any two UAVs \\ \hline
\( v_m \), \( h_m \) & Velocity of UAV \( m \) during flying and its altitude &
\( P_0, P_i \) & Blade profile power and induced power in hovering \\ \hline
\( U_{\text{tip}} \), \( v_0 \), \( a \) & Speed, mean speed and disc area of UAV's rotor & \( d_0 \), \( s \), \( \rho \) & Fuselage drag ratio, rotor solidity and air density  \\ \hline
\( T \), \( I \) & Total time period and total number of time slots &
\( \tau \), \( c \) & Duration of each time slot and speed of light \\ \hline
\( L_{n,m}(t) \), \( a \)  & Path loss between IoTD and UAV and its exponent & \( f_c \) & Carrier frequency  \\ \hline
\( \eta^{\text{LoS}}, \eta^{\text{NLoS}} \) & Path loss coefficients for LoS and NLoS &
\( P_{n,m}^{\text{LoS}}(t) \) & Probability of LoS between IoTD \( n \) and UAV \( m \) \\ \hline
\( \mu_1, \mu_2 \) & Environmental factors for LOS probability & \( B_m \), \( B_0 \) & Bandwidth allocated to UAV and HAPS \\ \hline
\( p_{n}^t \) & Transmission power between IoTD \( n \) and UAV \( m \) &
\( \sigma_m^2 \), \( \sigma_h^2 \) & Noise power in the UAV and HAPS comm. model \\ \hline
\( R_{m,0}(t) \) & Uplink data rate between UAV \( m \) and HAPS & \( R_{n,0}(t) \) & Uplink data rate between IoTD \( n \) and HAPS \\ \hline
\(R_{n,m}(t) \) & Uplink data rate between IoTD \(n\) and UAV \(m\) & \( SINR_{m,0}(t)\) & SINR between UAV \(m\) and HAPS \\ \hline 
\(SINR_{n,0}(t) \) & SINR between IoTD \( n \) and HAPS & \(SINR_{n,m}(t) \) & SINR between IoTD \( n \) and UAV \(m\) \\ \hline
\( g_{n,m}^0(t) \) & Reference channel gain between IoTD \( n \) and UAV \(m\)  & \( g_{n,m}(t) \) & Channel gain between IoTD \( n \) and UAV \( m \)  \\ \hline
\( d_{m,0}(t) \) & Distance between UAV \( m \) and HAPS &
\( Z_n(t) \) & Task set for IoTD \( n \) during time slot \( t \) \\ \hline
\( j_n(t) \) & Task arrival at IoTD \( n \) measured in bits & \( s_n(t) \) & Number of CPU cycles required for computing one bit \\ \hline
\( t_n^{\max}(t) \) & Maximum allowable latency for IoTD \( n \)'s tasks & \( j_n^l(t) \) & Portion of tasks processed locally by IoTD \( n \) \\ \hline
\( \alpha^o_n(t) \)  & Offloading ratio &\( j_n^o(t) \) & Portion of tasks offloaded by IoTD \( n \) \\ \hline
\( Q^l_n(t) \) & Local task buffer for IoTD \( n \) &
\( g_n^l(t) \) & Amount of tasks processed locally by IoTD \( n \) \\ \hline
\( f_n \) & Local processing rate of IoTD \( n \) & \( Q^o_{n,m}(t) \) & Offloading buffer for IoTD \( n \) at ABS \( m \) \\ \hline
\( g_{n,m}^o(t) \) & Size of tasks offloaded to ABS \( m \) from IoTD \( n \) & \( d_{n,m}(t) \) & Distance between IoTD \(n\) and UAV \( m \)\\ \hline
\( Q^e_{n,m}(t) \) & Edge buffer for tasks offloaded from IoTD to ABS & \( g_{n,m}^e(t) \) & Volume of tasks processed by ABS \( m \) from IoTD \( n \) \\ \hline
\( f_{n,m}(t) \) & Processing capacity of ABS \( m \) for IoTD \( n \) & \( \lambda_{n,m,0}(t) \) & Indicator if UAV \( m \) relays tasks of IoTD to the HAPS \\ \hline
\( Q^o_{n,m,0}(t) \) & Relaying buffer & \( g_{n,m,0}^o(t) \) & Volume of relayed tasks \\ \hline
\( Q^e_{n,m,0}(t) \) & Relaying edge buffer & \( g_{n,m,0}^e(t) \) & Volume of processed related tasks \\ \hline
\( f_{n,m,0}(t) \) & Processing capacity of HAPS for relayed tasks & \(\mathcal{Q}_{max}^l\) & Upper bound long-term local queue delay \\ \hline
\(\mathcal{Q}_{max}^o\) & Upper bound long-term offloading Queue delay & \(\mathcal{Q}_{max}^e\) & Upper bound long-term edge queue delay  \\ \hline
\(t_{n,m}^{o}(t)\) & Transmission delay for the offloaded tasks & \(t_{n}^{l}(t)\), \(t_{n,m}^{e}(t)\) & Local and edge computing delay \\ \hline
\( t_{n,m,0}^{o}(t)\) & Transmission delay for the relayed tasks & \( t_{n,m,0}^{o}(t)\) & HAPS computing delay for computing relayed tasks \\ \hline
\(W_n\), \(W_m\) & Effective switching parameter of IoTD and ABS \(m\) & \( E_{n,m}^{o}(t)\) & Energy consumption for task offloading \\ \hline
\(E_{n}^{l}(t)\), \( E_{n,m}^{e}(t)\) & Energy consumption for local and edge computing & \(E_{n,m,0}^{o}(t)\) & Energy consumption for relaying tasks  \\ \hline
\(E_{n,m,0}^{e}(t)\) & Energy consumption in computing the relayed tasks & \(E^{traj}_m(t)\) & Energy consumption of UAV during flying and hovering
\\ \hline
\end{tabular}
\end{table*}
\subsection{Mobility Model of IoTDs}
At the beginning of each operational period \( T \), specifically at the initial time slot \( t = 0 \), all IoTDs are initialized with random positions within the designated area. These devices follow the Gaussian-Markov mobility model, which aptly simulates realistic random fluctuations in their movements. Considering the short duration of each time slot, it is a reasonable approximation to consider the positions of IoTDs to be static within a single time slot. This approach simplifies the model without a significant loss of accuracy.

For each time slot \( t \), the velocity \( v_n(t) \) and the direction \( \theta_n(t) \) of a specific IoTD \( n \) evolve according to the following dynamic equations:
\begin{subequations}
\begin{align}
    v_n(t) &= \omega_1 v_n(t - 1) + (1 - \omega_1) \bar{v} + \sqrt{1 - \omega_1^2} \Phi_n,  \\
    \theta_n(t) &= \omega_2 \theta_n(t - 1) + (1 - \omega_2) \bar{\theta} + \sqrt{1 - \omega_2^2} \Psi_n,
\end{align}
\end{subequations}
where \( 0 \leq \omega_1, \omega_2 \leq 1 \) are coefficients that quantify the dependency on the previous state, capturing the inertia and directional persistence of the IoTDs. Here, \( \bar{v} \) and \( \bar{\theta} \) represent the average velocity and direction of the IoTDs across the network, respectively. The terms \( \Phi_n \) and \( \Psi_n \) are independent Gaussian random variables with respective mean-variance pairs \((\tilde{\xi}_{v_n}, \sigma_{v_n}^2)\) and \((\tilde{\xi}_{\theta_n}, \sigma_{\theta_n}^2)\) for each IoTD \( n \). Specifically, \(\Phi_n\) has a mean \(\tilde{\xi}_{v_n}\) and variance \(\sigma_{v_n}^2\), while \(\Psi_n\) has a mean \(\tilde{\xi}_{\theta_n}\) and variance \(\sigma_{\theta_n}^2\). These parameters \((\tilde{\xi}_{v_n}, \sigma_{v_n}^2)\) and \((\tilde{\xi}_{\theta_n}, \sigma_{\theta_n}^2)\) define the statistical properties of the random variables affecting the velocity and direction of the IoTDs.

The updated position coordinates \( (x_n(t), y_n(t)) \) of each IoTD are computed as follows:
\begin{subequations}
\begin{align}
x_{n}(t) &= x_{n}(t - 1) + v_n(t - 1)   \cos\Big(\theta_n(t - 1)\Big) \tau, \\
y_{n}(t) &= y_{n}(t - 1) + v_n(t - 1)   \sin\Big(\theta_n(t - 1)\Big) \tau,
\end{align}
\end{subequations}
since $\tau$ represents the time slot duration. This model, through its balance of deterministic trends and stochastic variations, offers a robust framework for simulating the nuanced and often unpredictable movements of IoTDs in dynamic and complex environments.

\subsection{UAV Trajectory Model}
In this model, we consider the dynamic positioning of UAVs over discrete time slots to optimize coverage and connectivity in the designated area, particularly in challenging environments like hotspots, disaster zones, or temporarily crowded areas. The movement of each UAV is defined by its trajectory, which is a function of both the direction and the distance it travels in each time slot \( t \). The trajectory of each UAV \(m\) is defined by two parameters: the angle \( \theta_{m}(t) \) and the distance \( d_{m}(t) \). The angle \( \theta_{m}(t) \in [0, 2\pi] \) represents the UAV’s direction of flight. The distance \( d_{m}(t) \in [0, d_{\text{max}}] \) represents how far the UAV travels, with \( d_{\text{max}} \) ensuring that the UAV remains within the operational boundary.

Assuming the initial coordinates of UAV \( m \) are \([X_{m,0}, Y_{m,0}]\), the position of UAV \( m \) at time slot \( t \) can be recursively calculated \cite{SHI2024103371} using its initial position and the sum of displacements over each time slot up to \( t \), as shown below:
\begin{subequations}
\begin{align}
X_{m}(t) &= X_{m,0} + \sum_{i=1}^{t} d_{m}(i)\cos(\theta_{m}(i)), \\
Y_{m}(t) &= Y_{m,0} + \sum_{i=1}^{t} d_{m}(i)\sin(\theta_{m}(i)).
\end{align}
\end{subequations}
Additionally, to ensure safe operation without collisions, the distance \( d_{m,m'}(t) \) between any two UAVs \( m \) and \( m' \) at any time slot \( t \) must be at least \( d_{\min} \). This safety constraint is crucial in preventing potential accidents and is given by:
\begin{equation}
d_{m,m'}(t) = \sqrt{\Big(X_{m}(t) - X_{m'}(t)\Big)^2 + \Big(Y_{m}(t) - Y_{m'}(t)\Big)^2}.
\end{equation}
Thus, we impose the condition:
\begin{equation}
d_{m,m'}(t) \geq d_{\min}, \quad \forall m, m' \in \mathcal{M}\setminus \{0\}, \quad m \neq m'.
\end{equation}
The energy consumption of UAVs is another critical factor in our model. It includes the energy required for flying, which is influenced by both the distance traveled and the UAV's velocity, as well as the energy consumed during hovering when no significant displacement occurs. The propulsion energy for a UAV \( m \) during flight is modeled based on the physical dynamics of rotary-wing aircraft, as introduced by Zeng \textit{et al}. \cite{zeng2019energy}:
\begin{equation}
\begin{split}
E_m^{fly}(t) &= \left[\frac{1}{2}d_0\rho s a \|v_m\|^3 + P_0 \left( 1 + \frac{3\|v_m\|^3}{U_{\text{tip}}^2}\right) \right. \\
& + P_i\left(\sqrt{1 + \frac{\|v_m\|^4}{4v_0^4}} - \frac{\|v_m\|^2}{2v_0^2}\right)^{\frac{1}{2}} \bigg] \frac{d_m(t)}{\|v_m\|},
\end{split}
\end{equation}
where \( P_0 \) and \( P_i \) are the blade profile power and induced power in hovering status, respectively; \( U_{\text{tip}} \) is the tip speed of the rotor blade; \( v_0 \) is the mean rotor velocity; \( d_0 \) is the fuselage drag ratio; \( s \) is the rotor solidity; \( \rho \) is the air density, and \( a \) is the rotor disc area.

For hovering, the energy consumed is proportional to the time the UAV remains stationary in the air, subtracting any time spent moving, as follows:
\begin{equation}
E_m^{\text{hov}}(t) = \left(P_0 + P_i\right) \left(\tau - \frac{d_m(t)}{\|v_m\|}\right),
\end{equation}
where \( \tau \) represents the duration of the time slot.
\subsection{User Association}\label{UA}
To manage the connectivity constraints of the UAVs, we define \(N^{max}_m\) as the maximum number of IoTDs each UAV \(m\) can serve. We use the binary indicator \(\beta_{n,m}(t)\), which equals \(1\) if IoTD \(n\) is served by UAV \(m\) at time \(t\), and \(0\) otherwise. This setup leads to the following constraints:
\begin{subequations}
\begin{align}
\beta_{n,m} &\in \{0,1\}, \quad \forall n \in \mathcal{N} , \forall m \in \mathcal{M}, \label{eq:sub00} \\
\sum_{n=1}^{N} \beta_{n,m}(t) &\leq N^{max}_m, \quad \forall m \in \mathcal{M} \setminus \{0\}, \label{eq:sub01}
\end{align}
\end{subequations}
ensuring that the number of IoTDs served by any UAV does not exceed their respective capacity limits. Additionally, each IoTD is associated with exactly one UAV or the HAPS, as expressed by:
\begin{equation}\label{eq1}
\sum_{m=0}^{M} \beta_{n,m}(t) = 1, \quad \forall n \in \mathcal{N}.
\end{equation}

In our adaptive offloading mechanism, IoTDs with tighter task delay constraints are prioritized for processing at the UAV level, where latency-sensitive tasks can be handled more rapidly. If the required offloading tasks exceed the UAV’s maximum computing capabilities or if a UAV \(m\) is at capacity, serving more IoTDs than \(N^{max}_m\), tasks from devices with less stringent delay constraints are offloaded to the HAPS. This adaptive multi-tier computing approach ensures that the network efficiently manages both real-time, delay-sensitive tasks and less urgent tasks by dynamically adjusting tasks offloading between UAVs and the HAPS.
\subsection{Fairness Model}
To ensure equitable service distribution among IoTDs and to encourage UAVs to serve diverse hotspots \cite{8713514} without clustering in the same areas, we introduce a hotspot fairness metric, \(f^{e}(t)\), tailored to our UAV service model. This metric extends the traditional concept of geographical fairness by incorporating hotspot-based service evaluation. This approach aligns with our objective to optimize UAV trajectories for balanced coverage across different geographical hotspots, ensuring that no single area is disproportionately served while others are neglected.

The hotspot fairness metric \(f^{e}(t)\) is defined to evaluate the fairness of UAV service distribution across these hotspots by tracking the number of IoTDs served by each UAV within the vicinity of each hotspot over time \cite{9209079}. This method allows us to assess not only the total number of IoTDs served but also the spatial distribution of these services relative to key areas of interest. We define the adapted hotspot fairness metric \(f^{e}(t)\) as follows:
\begin{equation}
f^{m}(t) = \frac{\left( \sum_{h=1}^{H} \sum_{i=1}^{t} N_{h,m}(i) \right)^2}{H \sum_{h=1}^{H} \left( \sum_{i=1}^{t} N_{h,m}(i) \right)^2},
\end{equation}
where \(N_{h,m}(i)\) indicates the total number of IoTDs from hotspot \(h\) served by UAV \(m\) at time slot \(i\). \(H\) represents the total number of hotspots. This metric reflects the level of fairness in distributing UAV services across different hotspots. A value closer to 1 indicates a more balanced distribution, suggesting that UAVs are effectively serving a diverse set of hotspots without undue concentration in specific areas. 
\subsection{Communication Model}
The communication model adopts the time division multiple access (TDMA) scheme by dividing the total time period \( T \) into \( I \) time slots, each of length \( \tau = \frac{T}{I} \). The set of the time slots is $\mathcal{I}=\{1,2,\dots, I\}$. Given the quasi-static nature of the system, the environment remains stable within each time slot \( \tau \) but may vary between slots.
\subsubsection{Uplink data rate from the IoTDs to the UAVs }
The data rate \( R_{n,m}(t) \) between the IoTD \( n \) and the UAV \( m \) during time slot \( t \) depends on the path loss and interference conditions. The path loss \( L_{n,m}(t) \) is composed of free-space path loss and the additional path losses incurred by Line-of-Sight (LoS) and Non-Line-of-Sight (NLoS)  links. The path loss is expressed as \cite{qin2022optimal}, \cite{article123}:
\begin{equation} \label{pathloss1}
\begin{split}
L_{n,m}(t) = & 20 \log_{10} \left(\frac{4\pi f_c d_{n,m}(t)}{c} \right) + P_{n,m}^{\text{LoS}}(t) \eta^{\text{LoS}} \\ 
& + \Big(1 - P_{n,m}^{\text{LoS}}(t) \Big) \eta^{\text{NLoS}},
\end{split}
\end{equation}
where \( d_{n,m}(t) = \sqrt{\| q_m(t)-q_n(t)\|^2 + h_m^2} \) represents the distance between IoTD \( n \) and ABS \( m \). Here, \( f_c \) is the carrier frequency, \( c \) is the speed of light, and \( h_m \) is the fixed altitude of UAVs. \( \eta^{\text{LoS}} \) and \( \eta^{\text{NLoS}} \) are the excessive path loss coefficients for LoS and NLoS losses, respectively \cite{9121255}.
The LoS probability \( P_{n,m}^{\text{LoS}}(t) \) is \{\cite{cheng2019space}:
\begin{equation}
P_{n,m}^{\text{LoS}}(t) = \frac{1}{1 + \mu_1\exp\left(-\mu_2\left[ \arctan\left(\frac{h_m}{d_{n,m}(t)}\right) -\mu_1\right]\right)},
\end{equation}
where \( \mu_1 \) and \( \mu_2 \) adjust the LoS probability based on environmental factors. It is worth noting that the path losses for LoS and NLoS are influenced by the distance between IoTD $n$ and UAV $m$, as reflected in the calculation of \( P_{n,m}^{\text{LoS}}(t) \).
The achievable data rate \( R_{n,m}(t) \) is given by:
\begin{equation}
\begin{split}
&R_{n,m}(t) = \\ & B_m \log_2 \left(1 + \frac{p_{n}^t 10^{-\frac{L_{n,m}(t)}{10}}}{ \sum_{i=1,i \neq n}^{N}\beta_{i,m} \alpha^o_{n} p_{i,m} 10^{-\frac{L_{i,m}(t)}{10}}+\sigma_m^2}\right),
\end{split}
\end{equation} 
where \( B_m \) is the bandwidth available to ABS \( m \), \( p_{n}^t \) is the transmission power between IoTD \( n \) and ABS \( m \), and \( \sigma_m^2 \) is the noise power. The term \( 10^{-\frac{L_{n,m}(t)}{10}} \) represents the channel gain \( g_{n,m}(t) \), and the summation in the denominator accounts for the intra-cluster interference among the IoTDs served by the same UAV \( m \).
\subsubsection{Uplink data rate to the HAPS}
If no UAV is in the coverage area to serve an IoTD \(n\), it can communicate directly with the HAPS. The uplink data rate from the IoTD \( n \) associated with the HAPS at time slot \( t \) is:
\begin{equation}
R_{n,0}(t) = B_0 \log_2\Big(1 + \text{SINR}_{n,0}(t)\Big),
\end{equation}
where the signal-to-interference-plus-noise ratio (SINR) between the IoTD \( n \) and the HAPS is:
\begin{equation}
SINR_{n,0}(t)=\frac{p^t_{n,0}(t) g_{n,0}^0(t) d^{-a}_{n,0}(t)}{\sum_{i=1,i \neq n}^{N}\beta_{i,0}.\alpha^o_{n}.p_{n}^t(t) g_{i,0}^0(t) d^{-a}_{i,0}(t) + \sigma_h^2 },
\end{equation} 
where \( d_{n,0}(t) = \sqrt{\| q_0(t) - q_n(t)\|^2 + h_0^2} \) represents the distance between the IoTD \( n \) and the HAPS. Here, \( a \) is the path loss exponent, \( B_0 \) is the bandwidth allocated to the HAPS, \( g_{n,0}^0(t) \) is the channel gain at a reference distance of 1 meter, and \( \sigma_h^2 \) is the noise power.
\subsubsection{Uplink data rate from UAVs to the HAPS}
When the computational resources of UAV \(m\) are insufficient to handle the computational tasks from all associated IoTDs, these tasks are relayed to the HAPS for further processing. The uplink data rate from UAV \( m \) to the HAPS at time slot \( t \) is:
\begin{equation}
R_{m,0}(t) = B_0 \log_2(1 + \text{SINR}_{m,0}(t)),
\end{equation}
where the SINR between UAV \( m \) and the HAPS is:
\begin{equation}
\text{SINR}_{m,0}(t) = \frac{p_{m}^t   g_{m,0}^0(t)   d_{m,0}^{-a}(t)}{\sum_{i=1,i \neq m}^{M} \alpha^o_{m}   p_{i,0}(t)   g_{i,0}^0(t)   d_{i,0}^{-a}(t) + \sigma_h^2},
\end{equation}
where \( d_{m,0}(t) = \sqrt{\| q_m(t) - q_0(t)\|^2 + (h_m - h_0)^2} \) is the distance between UAV \( m \) and the HAPS \cite{lakew2022intelligent}. Here, \( a \) is the path loss exponent, \( B_0 \) is the bandwidth allocated to the HAPS, \( p_{m}^t \) is the transmission power from the UAV to the HAPS, \( g_{m,0}^0(t) \) is the channel gain at a reference distance of 1 meter, and \( \sigma_h^2 \) is the noise power. 

In our work, we consider the IoTDs-to-UAV, IoTDs-to-HAPS, and UAV-to-HAPS communication models that differ in terms of bandwidth allocation, transmission power, distance between communicating nodes, and noise power. These differences are reflected in the SINR and data rate calculations for each model.
\subsection{Adaptive Queue-Aware Task Offloading Model }\label{comp}
This subsection introduces the adaptive queue-aware task offloading model in the MAGIN framework. In this model, each IoTD \( n \) manages computation-intensive tasks, including real-time applications and Metaverse services, that need to be executed within strict time constraints during discrete time slots. Each task set for IoTD \( n \) during time slot \( t \) is denoted by \( Z_n(t) = \{j_n(t), s_n(t), t_n^{\max}(t)\} \). Here, \( j_n(t) \) represents the volume of tasks arriving at time \( t \), measured in bits. The parameter \( s_n(t) \) indicates the number of CPU cycles required per bit to complete the computation. Finally, \( t_n^{\max}(t) \) specifies the maximum allowable latency for the tasks. Given the limited computational and energy resources of IoTDs, efficient task handling is essential. This is particularly important for demanding tasks that exceed local processing capabilities due to stringent QoS requirements, such as tight latency constraints. Once the user association is determined, each IoTD \( n \) will process a portion of its computational tasks locally while offloading the remainder to its associated ABS \( m \), which could be any UAVs or the HAPS. This model dynamically adapts task offloading decisions based on the real-time state of the task queues, the latency constraints of tasks, and the varying computational capacities of the IoTDs, UAVs, and HAPS.

The portion of tasks \( j_n(t) \) processed locally by IoTD \( n \) during time slot \( t \) is given by:
\begin{equation}
j_n^l(t) = \Big(1 - \alpha^o_n(t)\Big)   j_n(t),
\end{equation}
where \( (1 - \alpha^o_n(t)) \) represents the fraction of the total tasks \( j_n(t) \) that remains on the IoTD \(n\). Conversely, the part of the tasks offloaded to the ABS \( m \) is:
\begin{equation}
j_n^o(t) = \alpha^o_n(t)   j_n(t),
\end{equation}
where \( \alpha^o_n(t) \in [0,1] \) is the offloading parameter that determines the proportion of tasks \( j_n(t) \) to be processed by the edge server. The edge server located at the air network is responsible for processing all tasks offloaded from the IoTDs within the time slot duration \( \tau \). 
\subsubsection{Queuing model for direct task offloading}
This model captures the essential dynamics of task processing and buffering within IoTD and edge servers, ensuring efficient handling of computational tasks under various constraints and demands. In each time slot, IoTDs generate new tasks that arrive randomly and need to be processed. For each IoTD \( n \), the task \( Z_n(t) \) can be segmented into multiple parts, expressed as \( j_n(t) = k_n(t)   g(t) \), where \( k_n(t) = \{0, 1, 2, \dots\} \) for all \( n \in \mathcal{N} \). This segmentation helps in managing tasks efficiently based on their computational requirements.

To accommodate these tasks, each IoTD \( n \) is equipped with two buffers: \( Q^l_n(t) \) to store tasks \( j_n^l(t) \) computed locally and \( Q^o_n(t) \) to store tasks \( j_n^o(t) \) designated for offloading to the edge network. The dynamics of these buffers are governed by:
\begin{equation}
   Q^l_n(t+1) = \max \left\{Q^l_n(t) - g_n^l(t), 0 \right\} + j_n^l(t),
\end{equation}
where \( g_n^l(t) \) represents the amount of tasks processed locally \cite{qin2022optimal} during the time slot \( t \). This is defined as:
\begin{equation}
   g_n^l(t) = \min \left\{Q^l_n(t), \frac{\tau   f_n}{s_n(t)}\right\},
\end{equation}
indicating the capacity of IoTD \( n \) to process tasks locally, where \( f_n \) is the local processing rate and \( \tau \) is the duration of the time slot. For offloaded tasks, the buffer update is:
\begin{equation}
   Q^o_{n,m}(t+1) = \max \left\{Q^o_{n,m}(t) - g_{n,m}^o(t), 0 \right\} + j_n^o(t),
\end{equation}
where \( g_{n,m}^o(t) \) is the size of tasks offloaded to the edge server of the ABS \( m \), calculated by:
\begin{equation}
     g_{n,m}^o(t) = \min \left\{Q^o_{n,m}(t), \sum_{m=0}^M \beta_{n,m}(t)   \tau   R_{n,m}(t) \right\},
\end{equation}
which represents the effective offloading capacity depending on the data rate \( R_{n,m}(t) \) of the link between the IoTD and the associated ABS m. At the edge server \( m \), there is a buffer \( Q^e_{n,m}(t) \) to store tasks offloaded from IoTD \( n \), updated as:
\begin{equation}
  Q^e_{n,m}(t+1) = \max \left\{Q^e_{n,m}(t) - g_{n,m}^e(t), 0 \right\} + g_{n,m}^o(t),
\end{equation}
where \( g_{n,m}^e(t) \) denotes the volume of tasks processed by the edge server, defined by:
\begin{equation}
     g_{n,m}^e(t) = \min \left\{Q^e_{n,m}(t), \sum_{m=0}^M \beta_{n,m}(t)   \frac{\tau   f_{n,m}(t)}{s_n(t)} \right\},
\end{equation}
indicating the processing power of the edge server for tasks from IoTD \( n \), determined by the specific allocated computing resources \( f_{n,m}(t) \) of the edge server for IoTD \( n \). 

\subsubsection{Queuing model for task relaying from UAVs to the HAPS}
When the computational resources of UAV \(m\) are insufficient to handle the computational tasks from all the associated IoTDs, the number of IoTDs covered by UAV \( m \) surpasses its capacity, the UAV \( m \) relays tasks from the surplus IoTDs to the HAPS for further processing.
We define \( \lambda_{n,m,0}(t) \), a binary indicator, as \( 1 \) if UAV \( m \) relays the tasks of IoTD \( n \) to the HAPS, and \( 0 \) otherwise. When \(\lambda_{n,m,0}(t) = 1 \), there are four queues involved in the task processing and relay process: the local and offloading queues at the IoTD \( n \), as described in the previous subsection, a relaying queue at UAV \( m \), and an edge queue at the HAPS. The relaying buffer is updated as follows:
\begin{equation}
\begin{split}
   &Q^o_{n,m,0}(t+1)=\\&\sum_{m=1}^M \lambda_{n,m,0}(t).\Big( \max \left\{Q^o_{n,m,0}(t) - g_{n,m,0}^o(t), 0 \right\} + Q^o_{n,m}(t)\Big),    
\end{split}
\end{equation}
where \( g_{n,m,0}^o(t) \) represents the volume of tasks relayed from UAV \( m \) to the HAPS, computed as:
\begin{equation}
g_{n,m,0}^o(t) =\sum_{m=1}^M \lambda_{n,m,0}(t).\min \left\{Q^o_{n,m,0}(t), \tau R_{m,0}(t) \right\}.
\end{equation}

At the HAPS, tasks are stored in a buffer \( Q^e_{n,m,0}(t) \), which accumulates tasks offloaded from IoTD \( n \) to UAV \( m \), and then relayed to the HAPS:
\begin{equation}
\begin{split}
     &Q^e_{n,m,0}(t+1) =\\&\sum_{m=1}^M \lambda_{n,m,0}(t).\max \left\{Q^e_{n,m,0}(t) - g_{n,m,0}^e(t), 0 \right\} + g_{n,m,0}^o(t), 
\end{split}
\end{equation}
where \( g_{n,m,0}^e(t) \) is the volume of tasks processed by the HAPS, defined as:
\begin{equation}
    g_{n,m,0}^e(t) = \sum_{m=1}^M \lambda_{n,m,0}(t).\min \left\{Q^e_{n,m,0}(t),   \frac{\tau   f_{n,m,0}(t)}{s_n(t)} \right\}.
\end{equation}
This equation determines the processing capacity of the HAPS for tasks relayed from IoTD \( n \) through UAV \( m \), where \( f_{n,m,0}(t) \) represents the processing rate of the HAPS.
\subsection{Queuing Delays}
The total task computing delay is divided into local computing, offloading, relaying, and edge computing delay \cite{anajemba2020optimal}. The local computing delay for IoTD \( n \) in time slot \( t \) is defined as:
\begin{equation}
t_{n}^{l}(t) = \min \left\{\frac{Q^l_{n}(t) s_n(t)}{f_n}, \tau \right\},
\end{equation}
where \( t_{n}^{l}(t) \) represents the time taken for IoTD \( n \) to execute the tasks buffered in the local queue. Here, \( f_n \) is the CPU frequency of IoTD \( n \).

For the tasks offloaded to the edge, the transmission delay for IoTD \( n \) offloading to the ABS \( m \) is given by:
\begin{equation}
t_{n,m}^{o}(t) = \min \left\{ \frac{Q^o_{n,m}(t)}{\sum_{m=0}^M \beta_{n,m}(t)   R_{n,m}(t)}, \tau \right\},
\end{equation}
where \( t_{n,m}^{o}(t) \) is the time required to transmit the offloaded tasks from IoTD \( n \) to the edge server hosted by ABS \( m \). The edge computing delay, which is the time for the ABS \( m \) to compute the offloaded tasks, can be expressed as:
\begin{equation}
t_{n,m}^{e}(t) = \min \left\{\frac{Q^e_{n,m}(t)   s_n(t)}{\sum_{m=0}^M \beta_{n,m}(t)   f_{n,m}(t)}, \tau \right\},
\end{equation}
where \( t_{n,m}^{e}(t) \) denotes the time taken by ABS \( m \) to process the tasks offloaded from IoTD \( n \). \( f_{n,m}(t) \) is the allocated computing resources from the ABS \( m \)  to execute the tasks offloaded from IoTD \( n \).

If the UAV relays tasks to the HAPS, an additional time cost, known as the relaying time, is incurred. This relaying time is the time taken to offload tasks from UAV \( m \) to the HAPS. The total offloading time \( t_{n,m,0}^{o}(t) \) is given by:
\begin{equation}
    t_{n,m,0}^{o}(t) =\min \left\{ \frac{Q^o_{n,m,0}(t)}{\sum_{m=1}^M \lambda_{n,m,0}(t)   R_{m,0}(t)}, \tau \right\}.
\end{equation}
The edge computing delay, \( t_{n,m,0}^{e}(t) \), for tasks that are relayed and then processed at the HAPS is defined as:
\begin{equation}
\begin{split}
t_{n,m,0}^{e}(t) = \min \left\{\frac{Q^e_{n,m,0}(t) s_n}{\sum_{m=1}^M \lambda_{n,m,0}(t)   f_{n,m,0}(t)}, \tau\right\}.
\end{split}
\end{equation}
The total task delay for IoTD \( n \), associated with ABS \( m \) at time slot \( t \) in this distributed computing system is calculated as follows:
\begin{equation}
\begin{split}\label{totaldelay}
    t_{n}(t) = \max \Big( & t_{n}^{l}(t), t_{n,m}^{o}(t) + t_{n,m,0}^{o}(t)  \\
    & + t_{n,m}^{e}(t) + t_{n,m,0}^{e}(t) \Big).
\end{split}
\end{equation}
According to Little's Law \cite{9500852}, the queuing delay is proportional to the ratio of the queue length to the task arrival rate. To analyze this, we define, first, the task arrival rate averaged by time for the local queue $\mathcal{Q}^l_n(t)$, the offloading queue $\mathcal{Q}^o_n(t)$ and the edge-computing queue $\mathcal{Q}^e_{n,m}(t)$ as: $\bar{j}_{n}^l(t) = \frac{1}{t} \sum_{i=0}^{t-1} j_{n}^l(i),\quad \bar{j}_{n,m}^o(t) = \frac{1}{t} \sum_{i=0}^{t-1} j_{n,m}^o(i),\quad \bar{g}_{n,m}^o(t) = \frac{1}{t} \sum_{i=0}^{t-1} g_{n,m}^o(i),$ respectively. 

The upper bounds for the long-term average queuing delays are defined as:

\begin{subequations}
\begin{align}
 \bar{\mathcal{Q}}^l_n(t) =\lim_{{T \rightarrow \infty}} & \frac{1}{T}\sum_{t=1}^{T}\frac{\mathcal{Q}^l_n(t)}{\bar{j}_{n}^l(t)} \leq \mathcal{Q}_{max}^l,\label{eq:sub21} \\ 
 \bar{\mathcal{Q}}^o_{n,m}(t) = \lim_{{T \rightarrow \infty}} & \frac{1}{T}\sum_{t=1}^{T}\frac{\mathcal{Q}^o_{n,m}(t)}{\bar{j}_{n}^o(t)} \leq \mathcal{Q}_{max}^o, \label{eq:sub22} \\ 
 \bar{\mathcal{Q}}^e_{n,m}(t) = \lim_{{T \rightarrow \infty}} & \frac{1}{T}\sum_{t=1}^{T}\frac{\mathcal{Q}^e_{n,m}(t)}{\bar{g}_{n,m}^o(t)} \leq \mathcal{Q}_{max}^e, \label{eq:sub23}
\end{align}
\end{subequations}
where $\mathcal{Q}_{max}^l$, $\mathcal{Q}_{max}^o$, and $ \mathcal{Q}_{max}^e$ are the upper bounds for the long-term average local computing, offloading, and edge computing queuing delays, respectively.
\subsection{Energy Consumption}
We can determine the communication and the computational energy consumption in the proposed network by summing the energy used for local computing $E_{n}^{l}(t)$, transmitting offloaded tasks $E_{n,m}^{o}(t)$, relaying tasks $E_{n,m,0}^{o}(t)$, edge computing $E_{n,m}^{e}(t)$, and computing relayed tasks $E_{n,m,o}^{e}(t)$. The energy consumption for local computing can be calculated as follows:
\begin{equation}
    E_{n}^{l}(t) = W_n f_n^3 t_{n}^{l}(t),
\end{equation}
where $W_n$ is the effective switching capacitance parameter \cite{6giot}, and it is related to the hardware capabilities of IoTD $n$. The energy consumption for transmitting the offloaded tasks to ABS m and for computing them at the edge servers can be expressed as:
\begin{equation}
E_{n,m}^{o}(t) = p^t_n t_{n,m}^{o}(t), 
\end{equation}
\begin{equation}
    E_{n,m}^{e}(t) = W_m t_{n,m}^{e}(t) \sum_{m=0}^{M}\beta_{n,m}(t) f^3_{n,m}(t),
\end{equation}
respectively. $W_m$ is the effective switching capacitance parameter of ABS$_m \forall m\in\mathcal{M}$. For the relayed tasks, the energy consumption for the task relaying from the UAV \(m\) to the HAPS and for the computing of these relayed tasks are:
\begin{equation}
E_{n,m,0}^{o}(t) = p^t_m t_{n,m,0}^{o}(t),   
\end{equation}
\begin{equation}
    E_{n,m,0}^{e}(t) = W_m t_{n,m,0}^{e}(t) \sum_{m=0}^{M}\lambda_{n,m,0}(t) f^3_{n,m,0}(t),
\end{equation}
respectively. We can define the total communication and computational energy consumption for IoTD \( n \) as:
\begin{equation}
    E^{com2}_n(t) = E_{n}^{l}(t) +  E_{n,m}^{o}(t) + E_{n,m}^{e}(t) + E_{n,m,0}^{o}(t) + E_{n,m,0}^{e}(t).
\end{equation}
In addition, the flying and hovering energy consumption of UAV \(m\) during time slot \(t\) is defined as:
\begin{equation}
    E^{traj}_m(t) = E_{m}^{fly}(t) + E_{m}^{hov}(t).
\end{equation}
The total energy consumption in the proposed multi-tier MAGIN network is calculated as:
\begin{equation}\label{energy}
  E^{all}(t) = \sum_{n=1}^{N} E^{com2}_n(t) + \omega \sum_{m=1}^{M} E^{traj}_m(t),
\end{equation}
where $\omega$ is a positive constant weight factor. 

Regarding the energy consumption of the HAPS, we assume that HAPS has sufficient power,  as commonly stated in the literature \cite{ren2022caching},\cite{qin2022optimal}. HAPS typically has larger power reserves than UAVs and IoTDs, so we focus on the energy constraints of UAVs and IoTDs, following prior works \cite{jia2022hierarchical}.
\subsection{Problem Formulation}
In this paper, we aim to minimize the total energy consumption in multi-tier MAGIN by jointly optimizing the offloading decisions, UAV trajectory, and computing resource allocation at time slots under the constraints of the maximum queue capacities as well as the computing resource capabilities of the edge servers. The optimization problem is expressed as $\pmb{P_1}$ since $\mathcal{A}=\{\alpha^o_{n}(t)\}$ is the offloading ratio vector, $\mathcal{F}=\{f_{n,m}(t),f_{n,m,0}(t)\}$ is the allocated computing resources to compute the offloaded and relayed tasks, and $Q= \{q_{m}(t)\}$ is the UAV trajectory plan. Constraints \text{$C_1$} restrict the value and the upper bound of the association parameter and the upper bounds of the long-term average queuing delays. Constraint \text{$C_2$} indicates the value of the offloading ratio and the relaying indicator. The upper bound of the edge-allocated computing resources to compute the offloaded tasks is obtained by \text{$C_3$}. The battery capacity of each UAV exceeds the energy consumed for flying, hovering, and task relaying over the entire time period, as specified in constraint \( C_4 \). Similarly, the battery capacity of each IoTD surpasses the energy required for local computing and task offloading, as delineated in constraint \( C_5 \). The UAV trajectory constraints are shown in \text{$C_{6}$}-\text{$C_{8}$}. Constraint \text{$C_{9}$} states that the total time for computing the tasks in the network should not exceed the maximum allowable task delay.
\begin{align} \label{optp}
\pmb{(P_1):} & \quad  
\min_{\mathcal{A},\mathcal{F},Q} \quad \sum_{t=1}^{I}E^{all}(t)\\ \nonumber
\textrm{s.t.} \quad 
& C_1: (\ref{eq:sub00}),(\ref{eq:sub01}), (\ref{eq1}), (\ref{eq:sub21}),(\ref{eq:sub22}), (\ref{eq:sub23}), \\ \nonumber
& C_2: \alpha^o_{n}(t) \in [0,1], \lambda_{n,m,0}(t)\in\{0,1\}, \quad \forall n\in \mathcal{N} \\ \nonumber
& C_3: \sum_{n=1}^N \beta_{n,m}(t) f_{n,m}(t) \le f_{m}^{max}(t),\forall m\in \mathcal{M}, \forall t\in \mathcal{I}, \\ \nonumber
& C_4: \sum_{t=1}^{I}\Big(E_m^{traj}(t) + \sum_{n=1}^N E_{n,m,0}^{o}(t)\Big)\le E_m^{max}, \\ \nonumber
& \quad \quad \forall m\in\mathcal{M}\setminus \{0\},\\ \nonumber
& C_5: \sum_{t=1}^{I}\left(E_{n}^{l}(t) + \sum_{m=1}^{M}E_{n,m}^{o}(t) \right)\le E_{n}^{max}, \forall n\in \mathcal{N}, \\ \nonumber
& C_{6}: \|q_i(t)-q_j(t)\| \geq d_{min}, \quad \forall i,j\in\mathcal{M}\setminus \{0\}, \\ \nonumber
& C_7: X_{min} \leq x_m(t)\leq X_{max}, \quad \forall m\in\mathcal{M}\setminus \{0\}, \\ \nonumber
& C_8: Y_{min} \leq  y_m(t)\leq Y_{max}, \quad \forall m\in\mathcal{M}\setminus \{0\},  \\ \nonumber
& C_9: t_{n}(t)\le t_{n}^{max}, \quad \forall n\in \mathcal{N},\forall t\in \mathcal{I},  \nonumber
\end{align}

Solving Eq. (\ref{optp}) is challenging due to the mixed-integer nonlinear constraints and the interaction among multiple decision variables in the objective function and dynamic constraints. The nonlinearity arises because the resources allocated for edge computing are cubed in the energy calculation. The offloading ratio decision depends on the edge's available resources and the communication layer's load level. Additionally, the UAV trajectories are optimized based on the IoTDs' demand loads, leading to coupled network decisions. The dynamic nature of the arriving tasks and the varying priorities and requirements of the IoTDs introduce dynamic constraints to the optimization problem.

Furthermore, the problem should be solved efficiently in each time slot due to the dynamically generated tasks and the mobility of the IoTDs. Traditional iterative techniques like alternating optimization and genetic algorithms, though powerful, are computationally intensive \cite{9485089} and may not scale well for real-time applications \cite{9254093} due to the high complexity of the problem \cite{SHI2024103371}. To address these challenges, we propose an efficient solution using a MADRL approach, which is well-suited for managing the complexities of dynamically changing environments and the need for rapid online decision-making.
\section{A Multi-agent DRL-based Proposed Solution}
In this section, we present the proposed solution based on HMADRL, specifically using the MAPPO-BD algorithm. We begin by defining the multi-agent MDP components then, we outline the proposed centralized training and decentralized execution (CTDE) framework. A detailed analysis of the MAPPO-BD algorithm is provided, illustrating its application in solving our optimization problem.
\subsection{Multi-Agent MDP Elements Formulation}
The optimization problem presented in Eq.(\ref{optp}) is multi-dimensional and dynamically evolves, reflecting the complexities typical of real-world scenarios. Consequently, this problem is aptly modeled as a multi-agent MDP. An MDP generally consists of a global state space \(S\), a global action space \(A\), and a reward function \(R\). In a multi-agent context, each agent's view of the environment's state is often partially observable, a situation common in systems designed with privacy considerations and distributed architectures in mind.

Let \(O_{i}(t)\) represents the observation of agent \(i\) at the time slot $t$. The global state \(s_t\) of the environment can be constructed by aggregating these partial observations from all heterogeneous agents.

The global state space \(S\) and the action space \(A\) are defined as the Cartesian products of the observation spaces \(O_i\) and action spaces \(A_i\) of all agents as $S = O_1 \times O_2\times \dots \times O_K$ and $A = A_1 \times A_2 \times \dots \times A_K$, respectively, since $K=N+M+1$ represents the total number of all heterogeneous agents. These formulations \cite{lakew2022intelligent} capture the comprehensive state and action domains by considering the contributions of each agent. To mitigate the complexities inherent in the decision-making process and to seek near-optimal solutions, it is practical to decompose the overall policy concerning optimization variables into three distinct sub-policies. Within this multi-agent system, the three types of agents correspond to three specific policy paradigms, which are detailed as follows:
\subsubsection{IoTD agents}
Each IoTD dynamically optimizes the offloading ratio for its tasks within every time slot, taking into account the specific characteristics of the task and the queuing delays incurred during local computing and task offloading processes. These offloading decisions are carefully calibrated to minimize the data transmitted to the edge network, which conserves energy dedicated to offloading and ensures that offloading delays remain within acceptable limits. The index set of the agents of the IoTDs is $\mathcal{K}_1=\{1,2,\dots, N\}$.
\begin{itemize}
    \item \textbf{Observations:} Each IoTD maintains awareness of its task requirements defined by the triplet $j_n(t), s_n(t), t_n^{\max}(t)$ and the queuing delays for both local computing $Q^l_n(t)$, and task offloading $ Q^o_n(t)$. The state of each IoTD at time \(t\) can be described by:
    $O_n(t) = \{j_n(t), s_n(t), t_n^{\max}(t), Q^l_n(t), Q^o_n(t)\}.$
    \item \textbf{Action:} The action \(a_n(t)\) chosen by the agents of the IoTDs determines the offloading ratio \(\alpha^o_n(t)\) of their tasks: $A_n(t) = \alpha^o_n(t)$.
    \item \textbf{Reward:} The reward function for the IoTD agent is designed to capture both the goal of minimizing energy consumption and the penalties for not adhering to latency constraints. The reward function must also reflect the detailed energy consumption profile of each IoTD and its corresponding UAV. Thus, the reward is formulated as:
\begin{equation}\label{userreward}
    r_n(t) = - \Big( E^{com2}_n(t) + \omega \sum_{m=1}^{M}\beta_{n,m}(t) E^{traj}_m(t) + p_n(t)\Big),
    \end{equation}
    where \(p_n(t)\) denotes the penalty for delay violations, given by:
    \begin{equation}
    \begin{split}
    p_n(t) = & \psi_1 \text{ReLU}\left( \bar{\mathcal{Q}}^l_n(t) - \mathcal{Q}_{\max}^l \right) + \\ 
    & \psi_2   \text{ReLU}\left( \bar{\mathcal{Q}}^o_{n,m}(t) - \mathcal{Q}_{\max}^{o} \right),
    \end{split}
    \end{equation}
where \(\text{ReLU}(.)\) is the rectified linear unit function used to impose non-negativity on the penalties. The penalties are scaled by \(\psi_1\), \(\psi_2\), factors that adjust the importance of delay constraints in the reward function.
\end{itemize}
\subsubsection{UAV agents}
Given the task profiles of the IoTDs, their offloading ratios, priorities, and their proximity to the UAVs, the UAVs optimize their flight trajectories. Subsequently, they determine the computational resources allocated to process the offloaded tasks, considering the permissible edge queuing delays. The index set of the agents of the UAVs is $\mathcal{K}_2=\{N+1, N+2, \dots, N+M\}$.
\begin{itemize}
    \item \textbf{Observations:} Each UAV observes the locations of all UAVs, including itself. Additionally, the UAVs capture the task requirements and edge queuing delays of the closest $N_m^{\max}$ IoTDs. Also, the distances between the UAV and those IoTDs are obtained. The observation for each UAV \( m \) at time \( t \), \( O_m(t) \), is defined as follows: $O_m(t) = \{ d_{m,1}(t), d_{m,2}(t), \ldots, d_{m,N_{\max}}(t), q_1(t), q_2(t),\ldots,\\ q_M(t), \alpha^o_1(t), \alpha^o_2(t), \ldots,\alpha^o_{N_{\max}}(t), j_1(t), j_2(t),\ldots,\\ j_{N_{\max}}(t), s_1(t), s_2(t), \ldots, s_{N_{\max}}(t), t_1^{\max}(t), t_2^{\max}(t),\\ \ldots, t_{N_{\max}}^{\max}(t), Q^e_1(t), Q^e_2(t),\ldots, Q^e_{N_{\max}}(t)\}.$
    \item \textbf{Action:} At each time slot, based on the captured observations, the agents of the UAVs need to optimize their trajectories in order to serve their associated devices with the best QoS. Then, each UAV optimizes its allocated computing resources to those devices associated with it since $A_m(t)=\{f_{n,m}(t),q_m(t)\}$.
    \item \textbf{Reward:} The reward of each UAV agent should consider minimizing the energy consumption of both itself and the associated IoTDs. Also, the agent of each UAV should have a penalization process for colliding with the other UAVs, flying out of the studied coverage area \cite{qin2023multi}, and allocating inadequate computing resources that will lead to exceeding the allowable edge queuing delay. It is essential to recognize that UAVs are assigned additional rewards based on the number of users within their coverage area. This incentive structure is crucial for our system's functionality. In the absence of such rewards, UAVs may opt to position themselves away from the IoTDs. Thus, the UAVs are intentionally avoiding providing services. This avoidance strategy could lead all IoTDs to offload their tasks to the HAPS, given the absence of proximate UAVs. Consequently, UAVs could potentially maximize their rewards by reducing the energy consumption linked to these IoTDs.
    \begin{equation}
    \begin{split}\label{uavrew}
           r_m(t) =  &- \Big(\beta_{n,m}(t)\sum_{n=0}^{N} E_{n}^{com2}(t) + \omega E^{traj}_m(t) + p_m(t)\Big) \\&  + r_m^{guide}(t), \quad \quad \forall m \in \mathcal{M}\setminus\{0\},  
    \end{split}
    \end{equation}
since
\begin{equation}
p_m(t)= p^{delay}(t) + p^{fly out}(t) + P^{collision}(t),
\end{equation}
\begin{equation}
    p^{delay}(t) = \psi_3   \beta_{n,m}(t)\sum_{n=0}^{N} \text{ReLU}\left( \bar{\mathcal{Q}}^e_{n,m}(t) - \mathcal{Q}_{\max}^e \right),
\end{equation}
denotes the penalty for exceeding the edge queuing delay limit since \(\psi_3\) is a penalty factor \cite{10504534},
\begin{equation}
    p^{fly-out}(t) = \mu_o \left| q_m[n] - \text{clip}(q_m[n], 0, W) \right|,
\end{equation}
denotes the penalty when UAVs try to fly out of the square boundary with width $W$, and $\mu_o$ is a penalty factor.  It is assumed that the UAVs will stop at the boundary if they try to fly out of it, and thus $X_m[n] \leftarrow \text{clip}(X_m[n - 1]+d_m(t).\cos{\theta_m(t)}, 0, W)$ and $Y_m[n] \leftarrow \text{clip}(Y_m[n - 1]+d_m(t).\sin{\theta_m(t)}, 0, W)$.
\begin{equation}
p^{collision} = \mu_c \sum_{j=1}^{M} \min \left( \frac{\left| q_m[n] - q_j[n] \right| - d_{min}}{d_{min}}, 0 \right),
\end{equation}
is the penalty for disobeying the safety distance $d_{min}$ between UAVs, and $\mu_c$ is corresponding penalty factor.

The extra incentive for UAV \( m \) at time slot $t$ is defined as:
\begin{equation}
r^{guide}_m(t) = \mu_g   \frac{U_{\text{cov}}(q_m(t))}{N_m^{max}},
\end{equation}
where \( U_{\text{cov}}(q_m(t)) \) is calculated by:
\begin{equation}
U_{\text{cov}}(q_m(t)) = \sum_{n=1}^{N} \mathbf{1}\left( d_{m,n}\leq R_m\right),
\end{equation}
since \( \mathbf{1}( ) \) is the indicator function, \( R_m \) is the UAV coverage radius, \( \mu_g \) is the coverage reward factor.
\end{itemize}
\subsubsection{HAPS agent}
As a permanent aerial edge server, the HAPS is tasked with optimizing the allocation of computing resources. This involves considering the task profiles and offloading ratios of the IoTDs to reduce the overall energy consumption within the network.
\begin{itemize}
    \item \textbf{Observations:} The HAPS observes the task profiles, the offloading ratios, and the edge queuing latency of all the IoTDs. 
     $O_{M+1}(t) = \{ \alpha^o_1(t), \alpha^o_2(t), \ldots,  \alpha^o_{N}(t), j_1(t), j_2(t),\ldots,
    j_{N}(t), s_1(t),\\ s_2(t), \ldots, s_{N}(t),  t_1^{\max}(t), t_2^{\max}(t),\ldots, t_{N}^{\max}(t), Q^e_1(t),\\ Q^e_2(t), \ldots, Q^e_{N}(t)\}$. 
    \item \textbf{Action:} The agent of the HAPS aims to optimize the allocated computing resources to the associated IoTDs so $a_{M+1}(t)=\{f_{n,m,0}(t)\}$.
    \item \textbf{Reward:} The reward for the HAPS agent should focus on reducing the energy consumption of the connected IoTDs and include penalties for allocating insufficient computing resources, which would result in violating the edge queuing delay limits.
\begin{equation}\label{hapsrew}
    \begin{split}
    r_{M+1}(t+1) = - \sum_{n=1}^{N}\beta_{n,M+1} E^{com2}_n(t) - p_{M+1}(t),
    \end{split}
    \end{equation}
    since
    \begin{equation}
    p_{M+1}(t) = \psi_4 \sum_{n=0}^{N}\beta_{n,m+1}(t)\text{ReLU}( \bar{\mathcal{Q}}^e_{n,m}(t) - \mathcal{Q}_{max}^e), 
    \end{equation}
where $\psi_4$ is a weight penalty factor.
\end{itemize}
\subsection{MAPPO-Based Proposed JUTQORA in Multi-tier MAGIN }
\begin{figure}[h]
\includegraphics[width=3.4in, height=9.5cm]{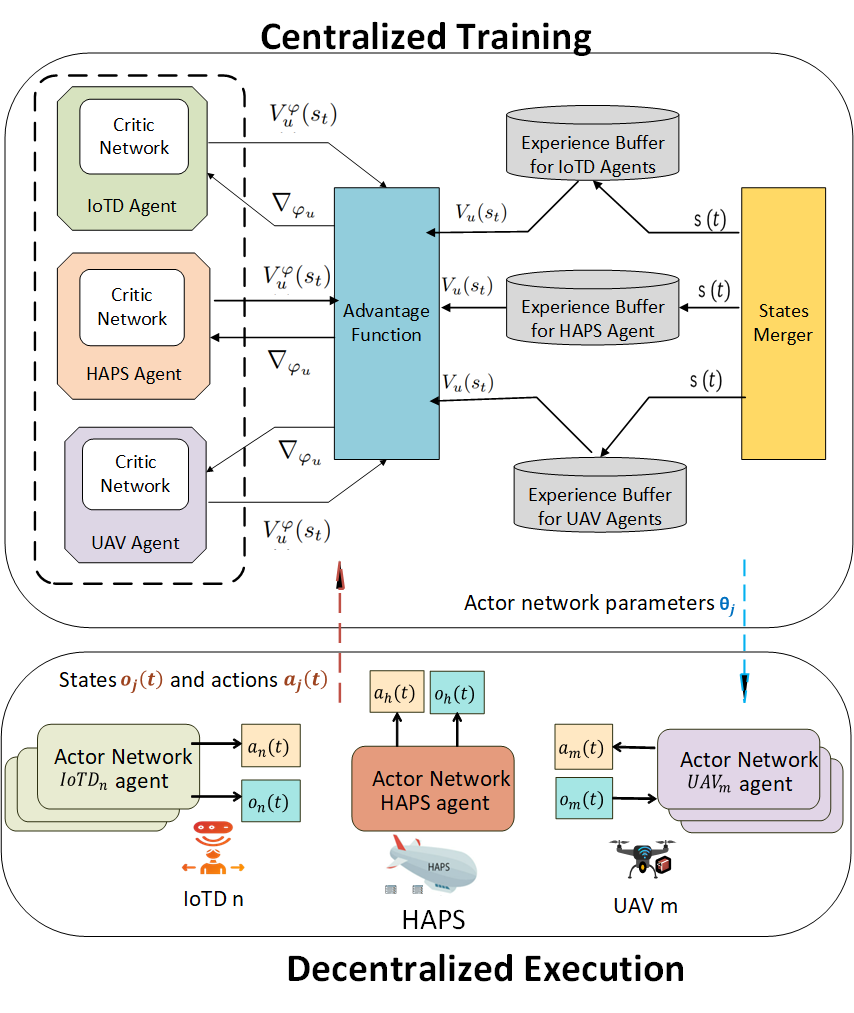}\centering
\caption{Proposed training framework of MAPPO-BD.} \label{fig:RLmodel}
\end{figure}

The MAPPO framework employs an actor-network, symbolized by \(\zeta_u\), to dictate the actions, while a critic network, denoted by \(\varphi_u\), assesses the state-value function of the agents with type $u$. This framework supports shared policies for the homogeneous agents, like IoTDs or the agents of the UAVs, represented by \(\pi_{\zeta_u}\). The architecture is designed to support straightforward deployment in distributed network environments by merging centralized training with decentralized execution, as illustrated in Fig. \ref{fig:RLmodel}.

In this structure, IoTDs, UAVs, and the HAPS undertake the task of offloading and computation resource allocation in accordance with the decisions from their actor networks. These actions are subsequently relayed to a central training hub, where the global environmental state is evaluated based on agent observations. Following this, the experience buffers are refreshed, and predictive values are derived. The networks, both actor and critic, are then updated, and the actor-network parameters are redistributed to IoTDs, the UAVs, and the HAPS. A key aspect of this framework is the sharing of network parameters among similar types of agents, which boosts efficiency and minimizes unnecessary duplication.

Within this framework \cite{10198525}, the state-value function for the \(u\)-th type of agent and the \(i\)-th agent is formulated as:
\begin{equation}
V^{\pi}_{u,i}(s_t, \zeta_u) = \mathbb{E}\left[\sum_{l=0}^{\infty} \gamma^l_u R_{u,i}(s_{t+l}, a_{t+l} \mid s_t = s, \pi)\right],
\end{equation}
where \(\mathbb{E}[.]\) signifies the expectation, \(R_{u,i}\) is the reward function for the \(i\)-th agent of the \(u\)-th type, \(a_t\) denotes the collective actions of all agents at time \(t\), \(s_t\) denotes the global state of all the agents at time \(t\), \(\pi\) represents the policy, and \(\gamma_u\) is the discount factor that signifies the relevance of future rewards. Similarly, the action-value function is articulated as follows:
\begin{equation}
\begin{split}
    Q^{\pi}_{u,i}&(s_t, a_t)  = \\ & \mathbb{E}\left[\sum_{l=0}^{\infty} \gamma^l_u R_{u,i}(s_{t+l}, a_{t+l} \mid s_t=s, a_t=a, \pi)\right].
\end{split}
\end{equation}
Based on the established framework, the advantage value of each action, essential for updating the policy, is represented by the advantage function \(A_{n,u,i} = Q^{\pi}_{u,i}(s_t, a_t) - V^{\pi}_{u,i}(s_t)\). This function can be effectively estimated using:
\begin{equation}
    \hat{A}_u(s_t) = \sum_{l=0}^{\infty} (\gamma_u \lambda)^l \left(r_{t+l} + \gamma_u V_u(s_{t+l+1}) - V_u(s_t)\right),
\end{equation}
where \(V_u(s_t)\) is the state-value function. It is important to highlight that the generalized advantage estimation (GAE), denoted by \(\lambda\), is utilized to approximate the advantage function. The GAE parameter \(\lambda\) plays a crucial role in managing the trade-off between bias and variance in the reward prediction.

Furthermore, the temporal-difference error is given by \(\delta_t = r_t + \gamma_u V_u(s_{t+1}) - V_u(s_t)\). Let \(V^\varphi_{u}(s_t)\) represent the state-value function as estimated by the critic network. The loss function for updating the critic network can then be expressed as:
\begin{equation}\label{vu}
   J(\varphi_u) = \frac{1}{2} \left( V^\varphi_{u}(s_t) - V_u(s_t) \right)^2. 
\end{equation}
For the actor networks, the clipping parameter \(\epsilon\) is integrated into the MAPPO algorithm to constrain the policy update ratio. The loss function for the actor-network is defined by:
\begin{equation}\label{ju}
    \begin{split}
     J(\zeta_u) = \mathbb{E}\left[\min \left( \text{clip}\left(\frac{\pi_{\zeta_u}(a_t|s_t)}{\pi_{\zeta'_u}(a_t|s_t)}, 1 - \epsilon, 1 + \epsilon \right), \right. \right. & \\ 
     \left. \left. \frac{\pi_{\zeta_u}(a_t|s_t)}{\pi_{\zeta'_u}(a_t|s_t)} \right) \hat{A}_u(s_t) \right] + \psi S_{t,u}, & 
    \end{split}
\end{equation}
where \(\zeta'_u\) are the parameters of the old policy. The ratio \(\frac{\pi_{\zeta_u}(a_t|s_t)}{\pi_{\zeta'_u}(a_n|s_t)}\) represents the update ratio, and \(\psi S_{t,u}\) denotes the policy entropy which encourages exploration.

The gradients \(\nabla_{\zeta_u} = \frac{\partial J(\zeta_u)}{\partial \zeta_u}\) and \(\nabla_{\varphi_u} = \frac{\partial J(\varphi_u)}{\partial \varphi_u}\) are used to update the parameters of the actor and critic networks respectively.
\subsection{MAPPO with Beta Distribution: MAPPO-BD}
In reinforcement learning, distributions in the actor-network are crucial for enabling exploration and defining probabilistic policies. The Gaussian distribution is traditionally used due to its simplicity and effectiveness in exploring continuous action spaces. However, it is unbounded, leading to challenges in environments where actions are naturally bounded. This often requires actions to be clipped, introducing boundary effects that can degrade learning performance.

To address these issues, Beta distribution is introduced as an alternative. With its support defined on the interval \([0, 1]\), which can be scaled to any \([a, b]\) interval, the Beta distribution aligns perfectly with bounded action spaces \cite{chou2017improving}. Its probability density function is given by:
\begin{equation}
   f(s; \alpha, \beta) = \frac{s^{\alpha - 1}(1 - s)^{\beta - 1}}{B(\alpha, \beta)}, 
\end{equation}
where \(\alpha\) and \(\beta\) are shape parameters, and \(B(\alpha, \beta)\) is the Beta function. This bounded support helps eliminate the need for action clipping, thereby preserving the integrity of the policy gradient.

\begin{algorithm}
\caption{Proposed MAPPO-BD-based JUTQORA Optimization algorithm in multi-tier MAGIN}
\begin{algorithmic}[1]
\STATE Initialize the maximum number of training episodes $epi_{max}$, episode length $I$, and the PPO epochs $epo_{max}$.
\STATE Initialize critic $\varphi_j$, and actor $\zeta_j$ networks for all agents of IoTDs, UAVs, and the HAPS $\forall$ \( j \in \{1,2,\ldots, K\}\).
\FOR{each training episode $\text{epi} = 1,2,\ldots,epi_{max}$}
    \FOR{each step $t = 1,2,\ldots,I$}
        \FOR{all the agents $j = 1,2,\ldots,K$}
            \STATE Obtain states $o_{j,t}$ and execute actions $a_{j,t}$.
            \STATE IoTD and UAV agents transmit their observations and actions to the HAPS.
            \STATE The HAPS computes the rewards of itself, the IoTDs, and the UAVs.
        \ENDFOR
        \STATE Obtain the log-probabilities for all the heterogeneous \\ agents \( u_{j,t} = \log \pi_{\zeta_j}(a_{j,t} | s_{j,t}) \), $\forall$ \( j \in \{1,2,\ldots, K\}\)
        \STATE Store the transitions $\{o_{j,t}, a_{j,t}, r_{j,t}, s_t, u_{j,t}\}$ in the memory buffers.
        \FOR{each epoch $\text{epo} = 1,2,\ldots,epo_{max}$}
            \FOR{each agent $j= 1,2,\ldots, K$}
                \STATE Update $\varphi_j$ and $\zeta_j$ using updates defined by equations \ref{ju} and \ref{vu}.
            \ENDFOR
        \ENDFOR
    \ENDFOR
\ENDFOR
\end{algorithmic}
\end{algorithm}

In the context of the MAPPO algorithm, adopting the Beta distribution is particularly advantageous for heterogeneous agents with varying action boundaries. It enables more uniform exploration and avoids the probability density concentration near the boundaries typical of Gaussian distributions. Consequently, MAPPO-Beta leads to improved learning dynamics and performance in multi-agent systems with continuous, diverse action spaces. We introduce the MAPPO-BD training framework, which is based on the Beta distribution. The pseudocode for this framework is provided in Algorithm~1.

\subsection{Complexity Analysis}
\label{sec:complexity}
The computational complexity of the proposed MAPPO-BD algorithm arises from the use of multi-layer perceptron (MLP) networks for both actor and critic components, as well as the interactions among multiple agents during centralized training and decentralized execution. In this section, we provide a detailed complexity analysis to address the scalability and efficiency of the proposed framework.

\subsubsection{Actor and Critic Network Complexity}
Each agent maintains separate actor and critic networks for decision-making and state evaluation, respectively. Since each agent may have a different number of layers and neurons in its networks, we denote $N_k$ as the number of layers for agent $k$, and $L_{n,k}$ as the number of neurons in the $n$-th layer of agent $k$. The computational complexity for a single forward pass through the MLP of agent $k$ is expressed as \cite{10198525}:
\begin{equation}
\mathcal{O}\left(\sum_{n=2}^{N_k-1} \left(L_{n-1,k} L_{n,k} + L_{n,k} L_{n+1,k}\right)\right),
\end{equation}
where $L_{n-1,k}$ and $L_{n+1,k}$ represent the number of neurons in the preceding and succeeding layers, respectively, for agent $k$. This complexity applies to both actor and critic networks.

\subsubsection{Complexity of Centralized Training and Decentralized Execution}
During centralized training, the actor and critic networks for all agents are updated iteratively using gradient descent. Assuming $\text{epi}_{\max}$ training episodes, $I$ time steps per episode, and $K$ agents, the total training complexity is given by:
\begin{equation}
\mathcal{O}\left(\text{epi}_{\max} \cdot I \cdot \sum_{k=1}^{K} \sum_{n=2}^{N_k-1} \left(L_{n-1,k} L_{n,k} + L_{n,k} L_{n+1,k}\right)\right).
\end{equation}
The centralized training process involves data collection, reward computation, and back-propagation for all agents, contributing to the overall complexity. Parallelization across agents mitigates the computational burden and improves training efficiency.

During decentralized execution, each agent performs a forward pass through its actor network to determine its optimal action. The per-step complexity for a single agent $k$ is:
\begin{equation}
\mathcal{O}\left(\sum_{n=2}^{N_k-1} \left(L_{n-1,k} L_{n,k} + L_{n,k} L_{n+1,k}\right)\right).
\end{equation}
This decentralized execution process is computationally efficient, as agents operate independently, ensuring suitability for real-time decision-making in dynamic network environments.

\subsubsection{Scalability and Efficiency}
The MAPPO-BD algorithm scales with the number of agents $K$ and the architecture of their respective actor and critic networks. Parallelization of agents and the use of shared policies for homogeneous agents (e.g., UAVs) improve scalability and reduce computational overhead. Additionally, the centralized training and decentralized execution framework ensures that the computational demands are distributed effectively across the system. While the training process is resource-intensive, it is performed offline, allowing the agents to learn optimal policies without real-time constraints. The execution process is lightweight, enabling efficient and scalable real-time decision-making in 6G-enabled air-ground integrated networks. This analysis demonstrates the feasibility of deploying MAPPO-BD in practical, large-scale scenarios.

\section{Evaluation Settings}
In this section, through an extensive simulation analysis, we demonstrate the effectiveness of the proposed MAPPO-BD-based JUTQORA in minimizing energy consumption while satisfying the problem constraints.
\subsection{Simulation Settings}
We investigated a coverage area of 
$1 \times 1$ Km serviced by a single HAPS and multiple UAVs. Each UAV functions as an auxiliary ABS and a flying edge server, capable of covering a circular area with a radius of $100$ meters and a maximum service capacity of $10$ IoTDs. The IoTDs are randomly distributed throughout the coverage area and uniformly within the hotspots. 
The upper bounds for the long-term average queuing delays are set at $140$ ms for local computing queues, $50$ ms for offloading queues, and $100$ ms for edge computing queues. The remaining parameter values of the proposed system \cite{lakew2022intelligent}, \cite{qin2022optimal} are summarized in Table \ref{tab:table2}.
For our experimental simulations, we utilized a Core i7 server featuring four cores, an Intel Xeon CPU operating at 2.3 GHz, and 16 GB of RAM. The experiments were conducted using Python 3.9.12.

To illustrate the effectiveness of our proposed MAPPO-BD algorithm, we conducted a comparative analysis against three benchmark schemes: \newline
\emph {1) PO-MAPPO-BD:} Proposed MAPPO-BD training algorithm with random UAV trajectory in the studied coverage area.\newline
\emph {2) MADDPG:} This algorithm is a popular method in MADRL. This off-policy algorithm features deterministic action outputs enhanced with exploration noise for effective learning. Each agent in MADDPG is equipped with dual actor networks and dual critic networks. The actor networks generate actions, and the critic networks evaluate these actions \cite{zhang2020uav}.\newline
\emph {3) MAPPO-ND:} This scheme employs a MAPPO-based training algorithm with a normal distribution applied to the actor-network \cite{kang2023cooperative}.
\begin{table}[h]
\centering \setlength{\abovecaptionskip}{0pt}
\setlength{\belowcaptionskip}{0pt}
\caption{SIMULATION PARAMETERS. } \label{tab:table2}
\begin{tabular}{|l|l|} \hline  
\textbf{Parameter}  & \textbf{Value} \\   \hline 
Altitude of HAPS $h_0$ and UAVs $h_m$ & 20 Km, 100 m \\    \hline  
Bandwidth of HAPS $B_{h}$ and UAVs $B_{m}$  & 100 MHz, $[20, 45]$ MHZ \\  \hline  
Noise power of HAPS $\sigma_h^2$ and UAVs $\sigma_m^2$ & -130 dBm, -144 dBm \\  \hline  
Additive loss of LoS link $\eta^{LOS}$ & 0.1 \\ \hline
Additive loss of NLoS link $\eta^{NLOS}$ & 21 \\ \hline
Environment Parameters $\mu_1, \mu_2$ & 4.88, 0.43 \\ \hline
Carrier frequency $f_c$ & 0.1 GHz \\ \hline   
Transmission power of IoTDs $p^t_n$ & $[20, 23]$ dBm \\ \hline  
Transmission power of UAVs $p^t_m$ & 30 dBm \\   \hline
Local computing resources $f_{n}$ & $[1, 2]$ GHZ \\ \hline   
Max. computing res. of UAVs $f_{n,m}^{max}$ & [18,20] \\  \hline 
Max. computing res. of HAPS $f_{n,0}^{max}$ & 100 GHZ \\  \hline 
Task size $j_n$ & [0.15, 0.45] MB \\   \hline  
Max. tolerable task delay $t_n^{max}$ & $[100, 500]$ ms \\   \hline   
CPU cycles for computing one bit $s_n$ & $[800, 1000]$ cycle/bit \\   \hline  
Effective switching capacitance $W_n$, $W_m$ & $10^{-28}$ \\   \hline
Number of IoTDs $N$ & $[50, 100]$  \\   \hline 
Number of UAVs $M$ and number of hotspots & $[2,5]$, $[4,7]$  \\   \hline 
UAV energy consumption weight factor $\omega$ & 0.001 \\ \hline
\end{tabular}
\end{table}
\subsection{Convergence Analysis}
In this subsection, we evaluated the convergence performance of the UAV path plan, task offloading, and computing resources allocation optimization problem. Achieving convergence in DRL is a challenging task due to the need for careful selection of learning parameters and the balancing of exploration versus exploitation, as well as the establishment of actor and critic networks. This challenge is further compounded in multi-agent heterogeneous DRL, where each agent's unique characteristics necessitate distinct parameter values. The parameter values of the MAPPO-based algorithms are organized in Table \ref{tab:table3}.
\begin{table}[ht]
\centering
\caption{PARAMETER SETTING OF MAPPO.}\label{tab:table3}
\begin{tabular}{|l|l|}
\hline
\textbf{Parameters} & \textbf{Value} \\ \hline
Total number of episodes $epi_{max}$ & 500  \\ \hline
Length of an episode $I$ & 50   \\ \hline
Learning rate of actor-network of IoTD and HAPS agents  & 0.001  \\ \hline
Learning rate of critic-network of IoTD and HAPS agents & 0.002  \\ \hline
Learning rate of actor-network of UAV agents  & 0.0001  \\ \hline
Learning rate of critic-network of UAV agents & 0.001  \\ \hline
Discount factor of IoTD, UAV, and HAPS agents $\gamma_u$ & 0.99  \\ \hline
Entropy bonus $\psi S_{t,u}$ and clipping parameter $\epsilon$  & 0.1, 0.2  \\ \hline
\end{tabular}
\end{table}

In this scenario, we have $70$ IoTDs generating tasks with sizes ranging from $0.25$ to $0.45$ MB, and the allowable delay spans from $100$ to $150$ ms. The remaining parameters are consistent with those presented in Table \ref{tab:table2}. 

\begin{figure*}[t]
    \centering
        \begin{subfigure}{0.33\textwidth}
        \includegraphics[width=\textwidth]{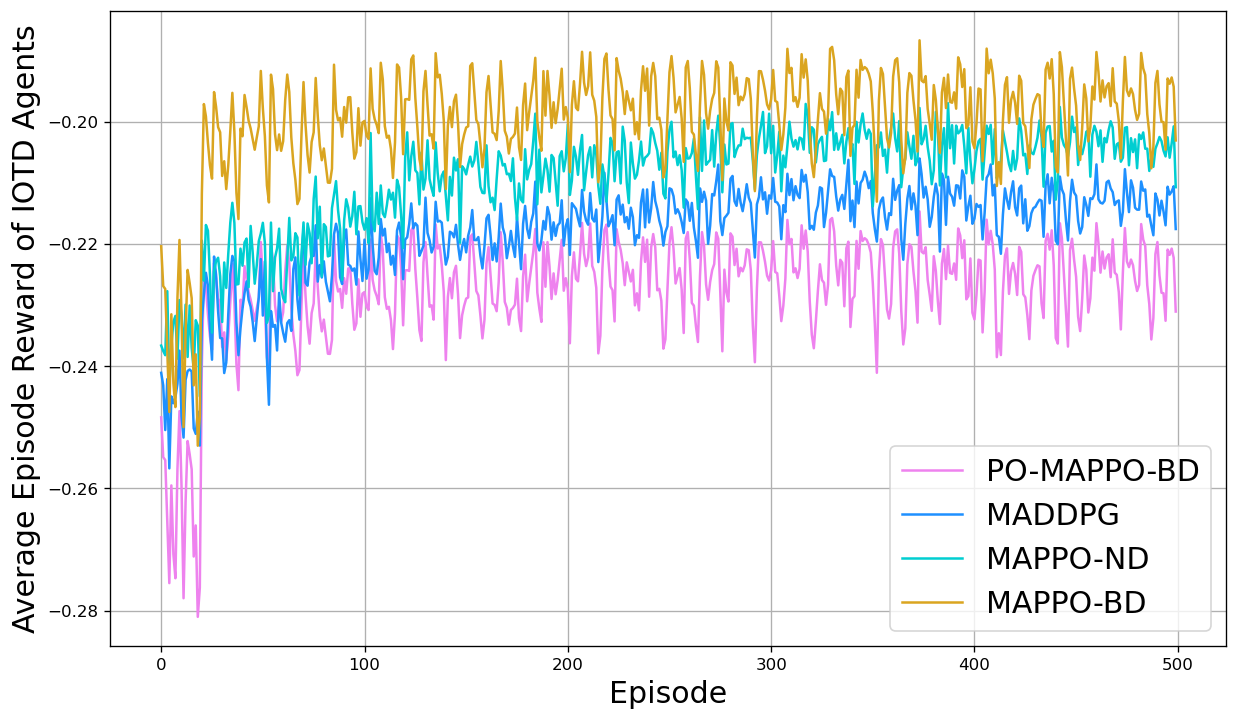}
         \caption{Average episode reward of IoTDs.}
        \label{fig: UserRew}
        \end{subfigure}
     \begin{subfigure}{0.33\textwidth}
        \includegraphics[width=\textwidth]{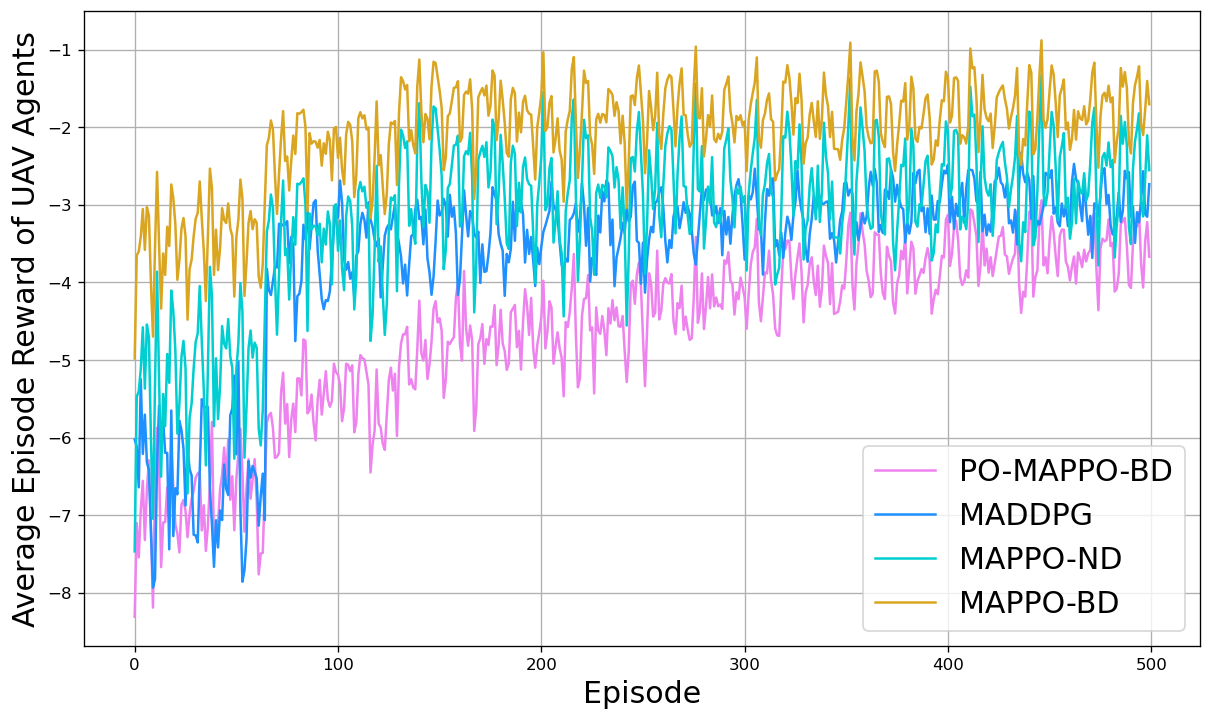}
         \caption{Average episode reward of UAVs.}
        \label{fig: UAVRew}
        \end{subfigure}
       \begin{subfigure}{0.33\textwidth}
    \includegraphics[width=\textwidth]{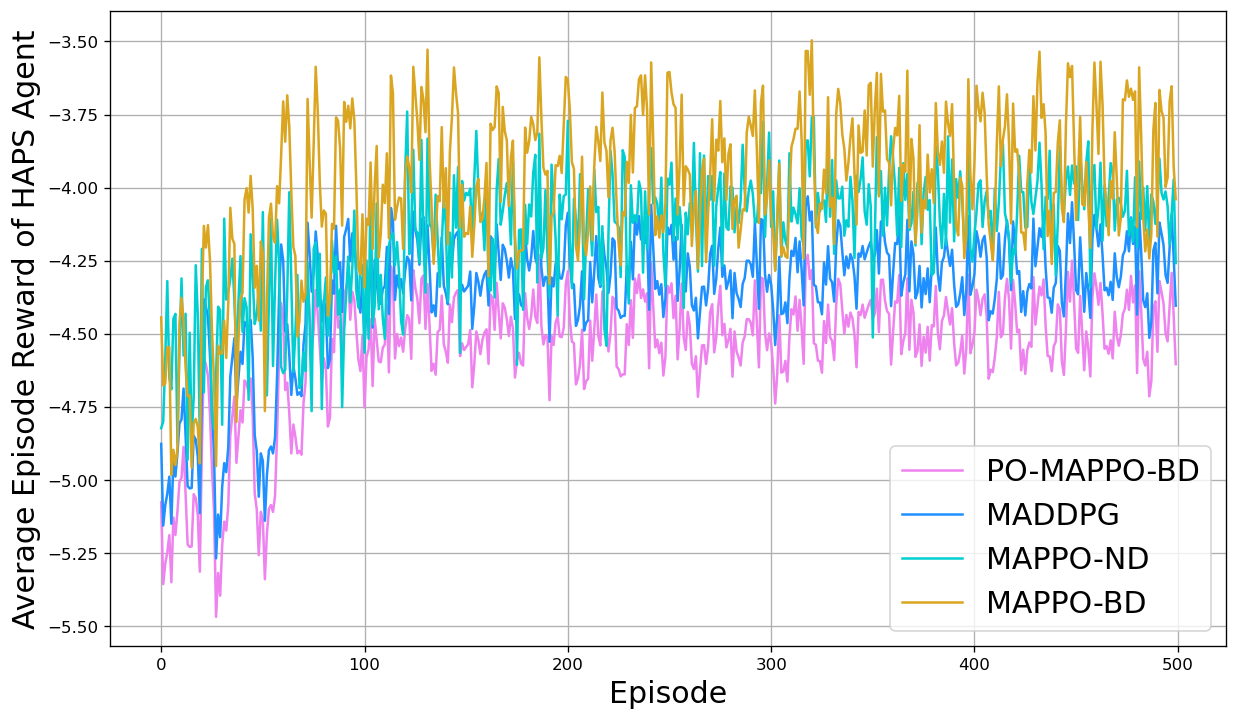}
    \caption{Average episode reward of HAPS.}
    \label{fig: HAPSRew}
    \end{subfigure}
    \caption{Reward vs. Episode Analysis.}
    \label{RewardVsEpi} 
\end{figure*}

Sub-figures \ref{fig: UserRew}, \ref{fig: UAVRew}, and \ref{fig: HAPSRew} present the average rewards of all IoTD agents, UAV agents, and the HAPS agent, respectively, within a single episode. These rewards are calculated according to Eq. (\ref{userreward}), Eq. (\ref{uavrew}), and Eq. (\ref{hapsrew}), respectively. The IoTDs optimize the offloading ratio of computing tasks based on their capabilities and the profiles of the tasks. The UAVs effectively position themselves in optimal locations and allocate appropriate computing resources to efficiently serve the IoTDs. The HAPS allocates computing resources to IoTDs outside the coverage of all UAVs or relayed by overloaded UAVs to the HAPS.
It is evident that our proposed MAPPO-BD algorithm outperforms the other baselines, achieving the highest reward values. The MAPPO-ND algorithm, which employs a MAPPO-based training scheme, surpasses the performance of both the MADDPG and PO-MAPPO approaches. The random UAV trajectory in the PO-MAPPO scheme leads to an unbalanced load distribution of IoTDs between the HAPS and UAVs, causing some UAVs to operate in areas without serving IoTDs. This behavior results in the poorest performance in terms of maximizing the reward of IoTD agents. This observation underscores the critical importance of strategic trajectory planning and effective resource management in achieving optimal performance. In comparison, while the MADDPG algorithm demonstrates reliability, it falls short in optimizing performance when contrasted with the MAPPO-based approaches. This discrepancy highlights the advantages of the MAPPO framework in handling complex, multi-agent scenarios more effectively. 

In Fig. \ref{RewardVsEpi}, the average rewards for IoTD agents, UAV agents, and the HAPS agent gradually increase over training episodes and eventually converge. This shows that the agents are effectively learning and adjusting their strategies, maximizing cumulative rewards through interactions with the environment. The convergence of rewards indicates that the agents have reached an optimal policy set, ensuring consistent decision-making performance. Based on the analysis, the proposed MAPPO-BD algorithm converges relatively early, at episode 200 for IoTDs and at episode 250 for UAVs and HAPS agents. In comparison, PO-MAPPO-BD converges at episode 300, while MADDPG shows convergence at episode 350. MAPPO-ND achieves convergence earlier, at episode 250, for all agents. This analysis highlights the efficiency of the MAPPO-BD in achieving faster and more stable convergence compared to the other benchmark methods, demonstrating its robustness in dynamic multi-agent environments.

\begin{figure}[h]
    \centering
    \begin{subfigure}[b]{0.24\textwidth}
    \centering
    \includegraphics[width=\textwidth, height=3cm]{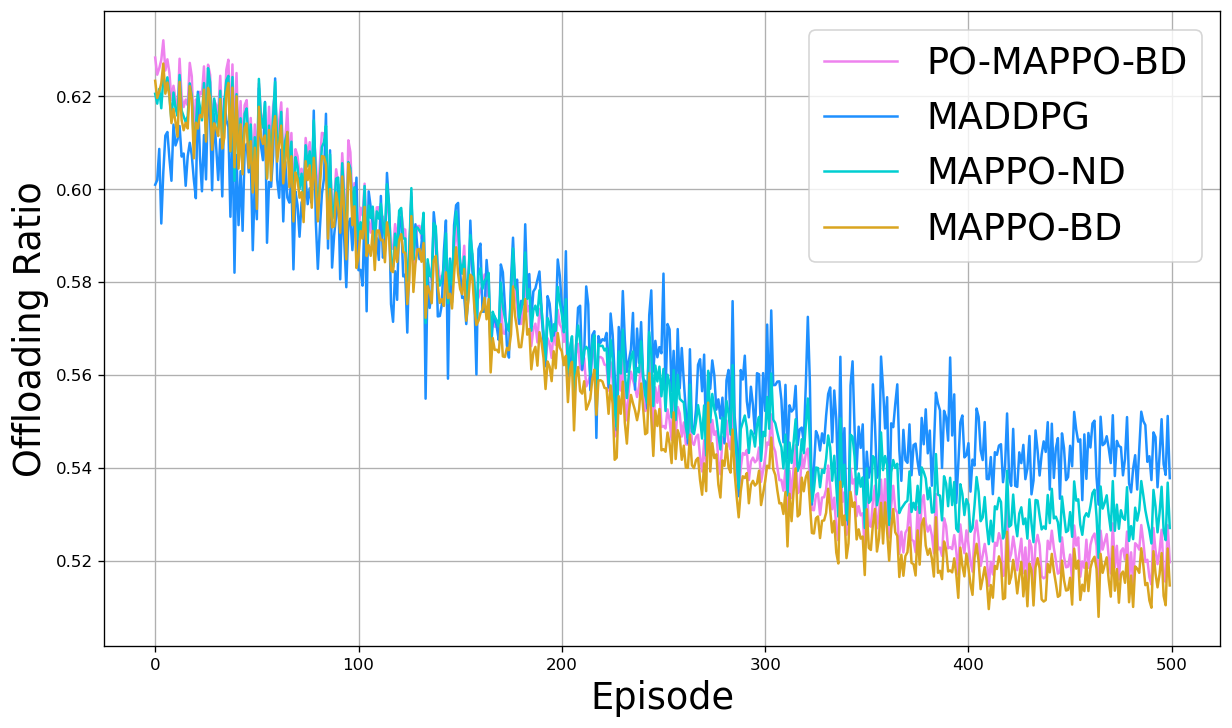}
    \caption{$\alpha_{n}^{o}$}
    \label{fig.offVsEpi}
     \end{subfigure}
    \hfill
    \begin{subfigure}[b]{0.24\textwidth}
    \centering
\includegraphics[width=\textwidth,height=3cm]{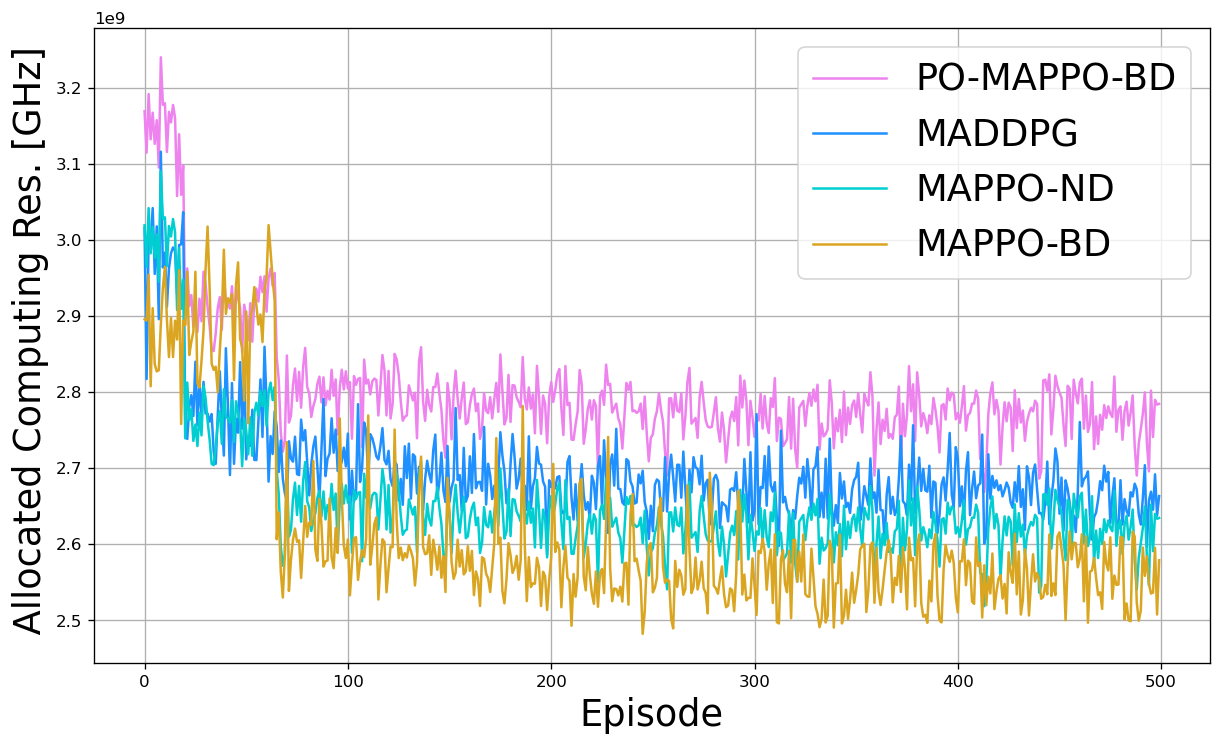}
    \caption{$f_{n}^{e}$}
    \label{fig.CompVsEpi}
    \end{subfigure}
    \hfill
    \begin{subfigure}[b]{0.24\textwidth}
    \centering
\includegraphics[width=\textwidth,height=3cm]{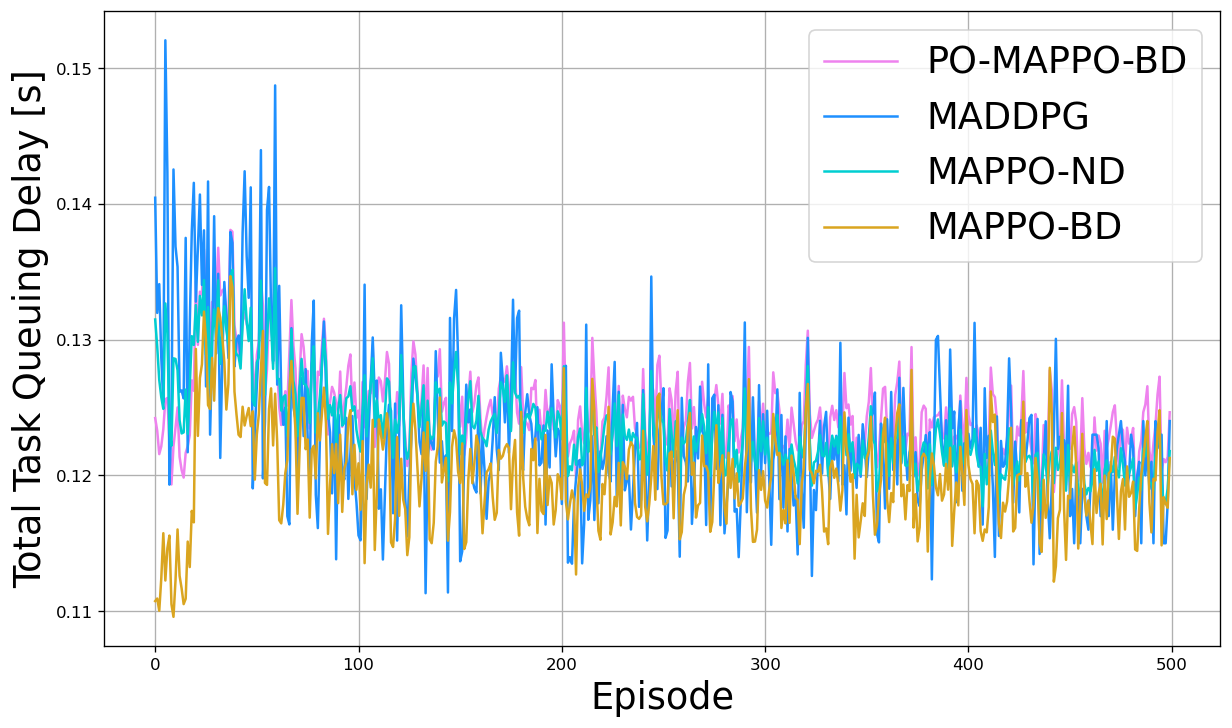}
    \caption{ $t_n$}
    \label{fig:TVsEpi}
    \end{subfigure}
    \hfill
    \begin{subfigure}[b]{0.24\textwidth}
    \centering
\includegraphics[width=\textwidth,height=3cm]{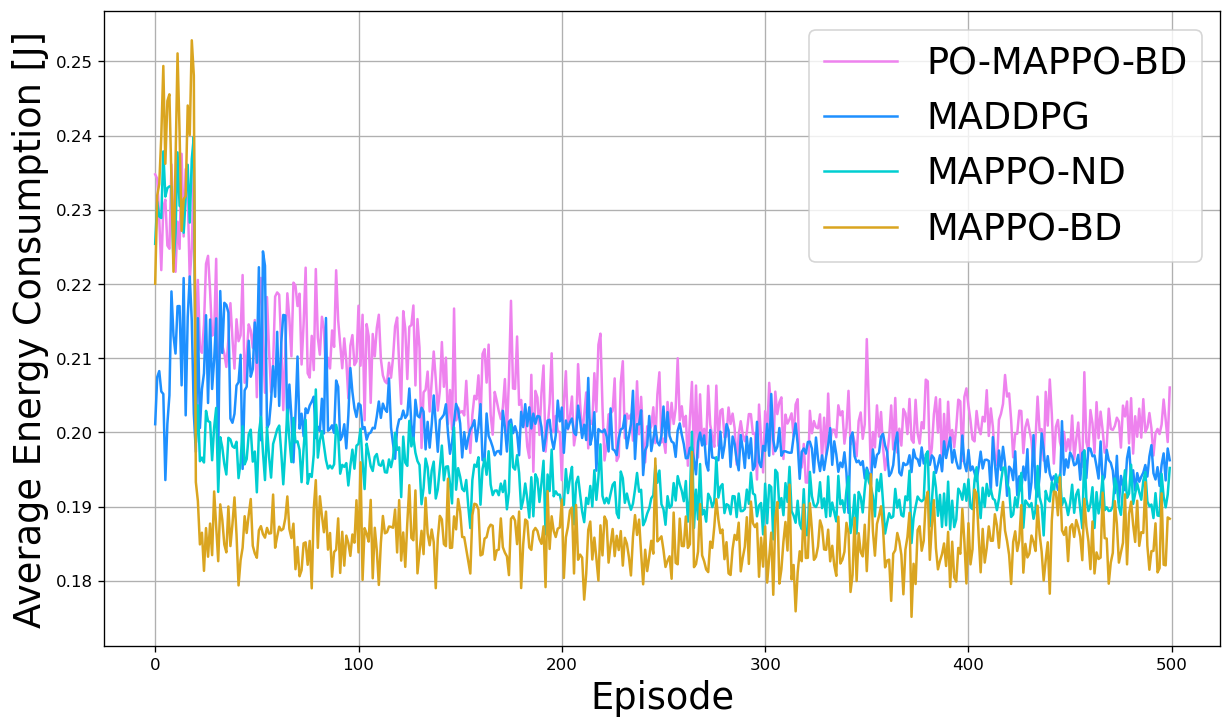}
    \caption{ $E^{all}$}
    \label{fig:EVsEpi}
    \end{subfigure} 
    \caption{ Convergence analysis.}
    \label{fig:convergence}
\end{figure}

The convergence analysis of action parameters related to task offloading and resource allocation optimization is shown in Fig. \ref{fig:convergence}. To meet the maximum task delay \( t_n^{max} \), IoTD \( n \) may offload some tasks to the associated ABS due to limited local computing capabilities. Figure \ref{fig.offVsEpi} illustrates the offloading ratio \( \alpha_n^o \). Using the MAPPO-BD algorithm, IoTD agents effectively utilize local resources, offloading about \( 51.9\% \) of their tasks, which is lower than other baseline methods. Offloading decisions depend on local computing and task profiles, not UAV trajectory optimization. This explains why PO-MAPPO-BD has similar offloading rates to MAPPO-BD. The MADDPG-based approach offloads the most tasks at \( 54.1\% \), likely due to its deterministic nature and limited adaptability to network dynamics. The MAPPO-ND method performs moderately, positioned between MADDPG and MAPPO-BD.

The computing resources allocated by UAVs and HAPS, denoted as \( f_n^e \), are averaged per episode and shown in Fig. \ref{fig.CompVsEpi}. Edge entities aim to minimize energy consumption by allocating the necessary computing resources while respecting maximum task delay. The MAPPO-BD algorithm stands out, using about 2.55 GHz, demonstrating effective resource management. The MAPPO-ND algorithm also performs well, surpassing the MADDPG and PO-MAPPO-BD baselines in resource allocation. However, PO-MAPPO-BD allocates the most resources at approximately 2.785 GHz, likely due to random UAV trajectories leading to inefficient edge resource utilization. While MADDPG performs better than PO-MAPPO-BD, it remains less efficient than MAPPO-ND and MAPPO-BD in computing resource allocation.

Fig. \ref{fig:TVsEpi} illustrates that all the approaches successfully maintain tight delay constraints, as the average total task latency \( t_n \) for all IoTDs in each episode remains below 125 ms. This value represents the midpoint of the maximum allowable delay range of 100 to 150 ms. The results demonstrate that all baseline algorithms MAPPO-BD, MAPPO-ND, MADDPG, and PO-MAPPO-BD effectively manage task delays, ensuring compliance with the specified latency constraints.

The primary goal of our optimization is to minimize total energy consumption in the proposed MAGIN. Our MAPPO-BD algorithm achieves the lowest average energy consumption of approximately 0.187 J, as shown in Fig. \ref{fig:EVsEpi}. In contrast, the PO-MAPPO-BD algorithm consumes around 0.202 J, highlighting the negative impact of random UAV trajectories on energy efficiency. Additionally, the MAPPO-ND algorithm outperforms the MADDPG-based approach, showcasing the advantages of probabilistic elements in reinforcement learning. Our algorithm minimizes task offloading to the edge, maximizing local resource use and reducing energy consumption. It allocates only the essential computing resources at the edge, further conserving energy. By optimally positioning UAVs among IoT devices needing to offload tasks, we promote effective resource use in multi-tier MAGIN environments. Fluctuations in the trends of the figures are noted due to the diverse capabilities of IoT devices and their dynamic task profiles. The values in Fig. \ref{RewardVsEpi} and Fig. \ref{fig:convergence} represent averaged data across all IoT devices, leading to observable variations in the plots. 

The stabilization of the results shown in Fig. \ref{fig:convergence} after sufficient training episodes demonstrates that the MAPPO-BD algorithm effectively learns and implements optimal policies for resource management and task offloading. This convergence ensures that the system can dynamically adapt to varying network conditions and application demands while maintaining optimal performance, which is critical for practical deployment in real-time environments.

\begin{figure}[h] 
\centering \includegraphics[width=3in, height=5.5cm]{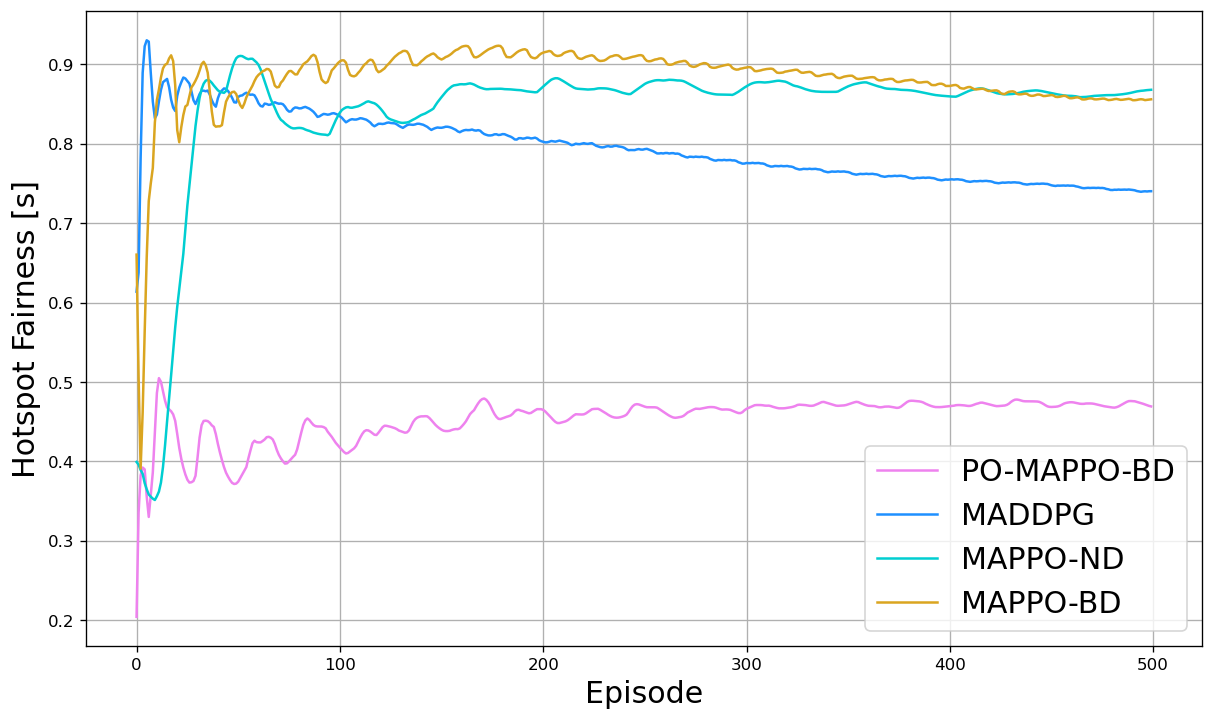} 
\caption{Hotspot Fairness.}
\label{fig: fair} 
\end{figure}

Fig. \ref{fig: fair} illustrates the hotspot fairness metric \( f^e(t) \) over 500 episodes comparing the proposed algorithm with the baseline methods. Although the fairness parameter is not explicitly an optimization parameter within the problem formulation, the dynamic nature of task requirements and arrivals from the IoTDs necessitates that the UAVs serve those IoTDs that achieve the highest energy savings. This requires the UAVs to determine which hotspot to serve in each episode, thereby maximizing energy savings and ensuring efficient service distribution. MAPPO-BD achieves the highest fairness values, around 0.86, indicating the most balanced service distribution among the algorithms evaluated. Given that there are five hotspots, a fairness value of 0.86 suggests that the UAVs are effectively covering four hotspots in each episode, ensuring they are not static and continuously moving to provide balanced service coverage. MAPPO-ND exhibits fairness values similar to MAPPO-BD, reflecting its capability to maintain a balanced service distribution. MADDPG starts with high fairness values comparable to MAPPO-BD and MAPPO-ND but gradually decreases to around 0.72. This decline can be attributed to the deterministic nature of MADDPG, where once the policy is learned, the UAVs' actions become predictable and fixed. While this can stabilize performance initially, it may lead the UAVs to favor specific hotspots consistently, resulting in an uneven service distribution and a declining fairness metric over time. In contrast, PO-MAPPO-BD achieves fairness values less than 0.5, indicating that during random trajectories, the UAVs only visit approximately two hotspots, demonstrating significant imbalance in service distribution. 
\subsection{Energy Consumption and Queuing Latency Analysis}
\begin{figure}[h]
    \centering
    \begin{subfigure}[b]{0.24\textwidth}
    \centering
    \includegraphics[width=\textwidth, height=3cm]{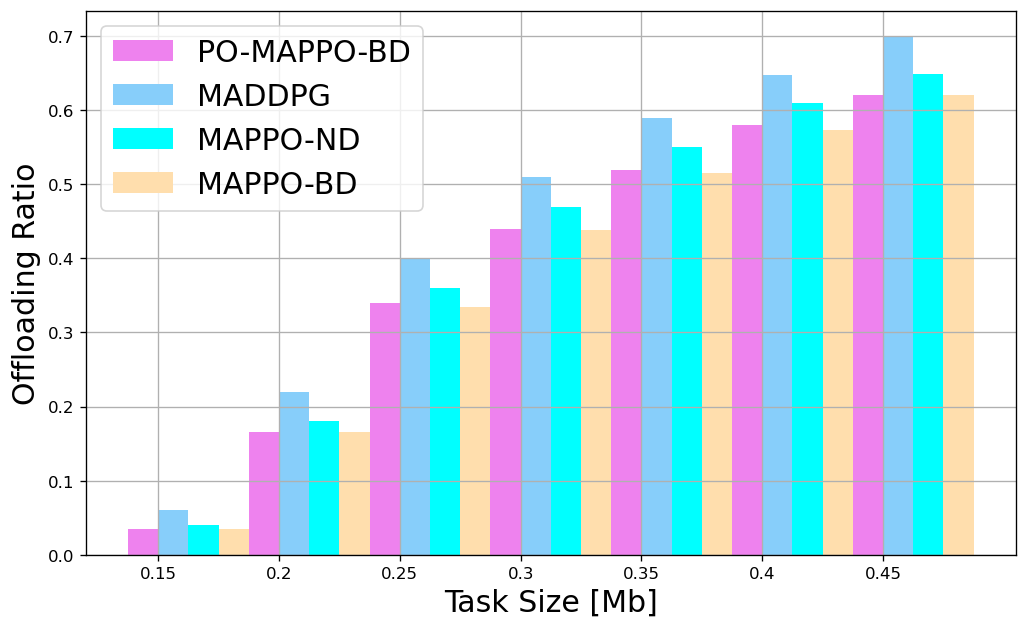}
    \caption{$\alpha_{n}^{o}$}
    \label{fig.offVsTs}
     \end{subfigure}
    \hfill
    \begin{subfigure}[b]{0.24\textwidth}
    \centering
\includegraphics[width=\textwidth,height=3cm]{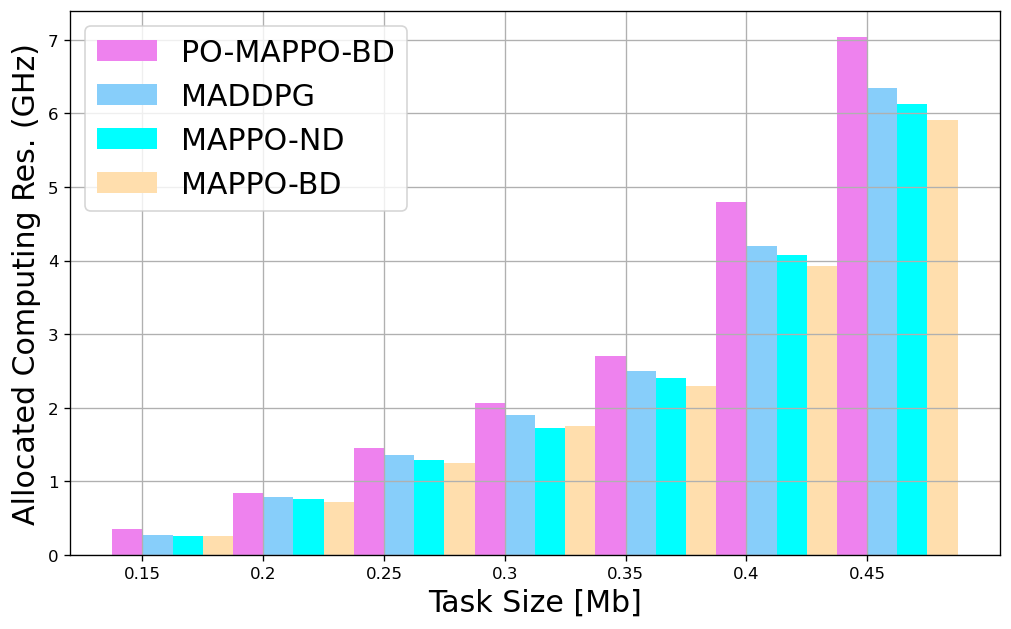}
    \caption{$f_{n}^{e}$}
    \label{fig.CompVsTs}
    \end{subfigure}
    \hfill
    \begin{subfigure}[b]{0.24\textwidth}
    \centering
\includegraphics[width=\textwidth,height=3cm]{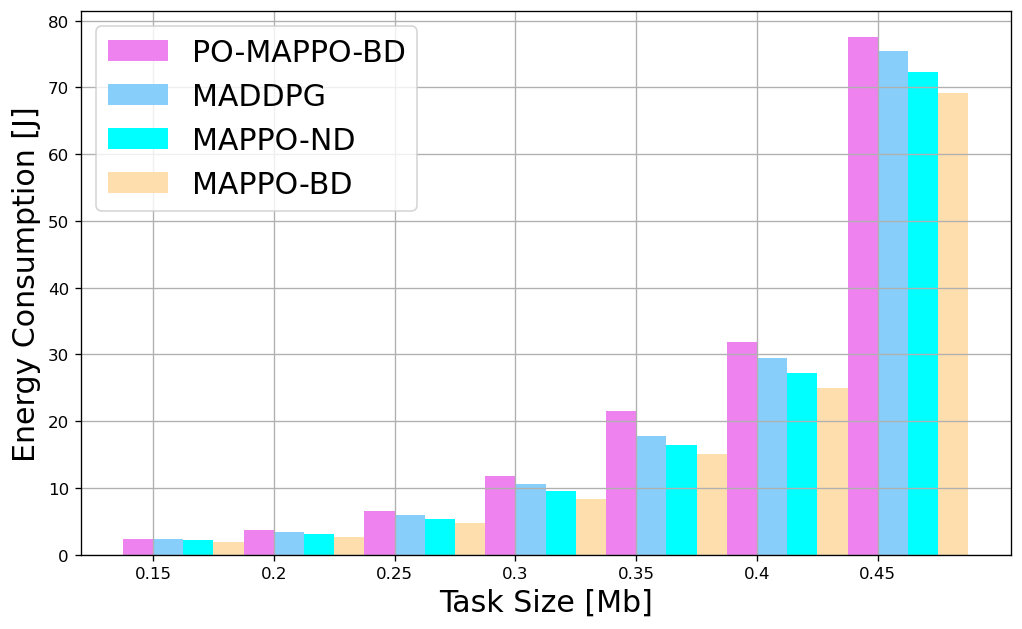}
    \caption{ $E^{all}$}
    \label{fig:EVsTs}
    \end{subfigure} 
    \hfill
    \begin{subfigure}[b]{0.24\textwidth}
    \centering
\includegraphics[width=\textwidth,height=3cm]{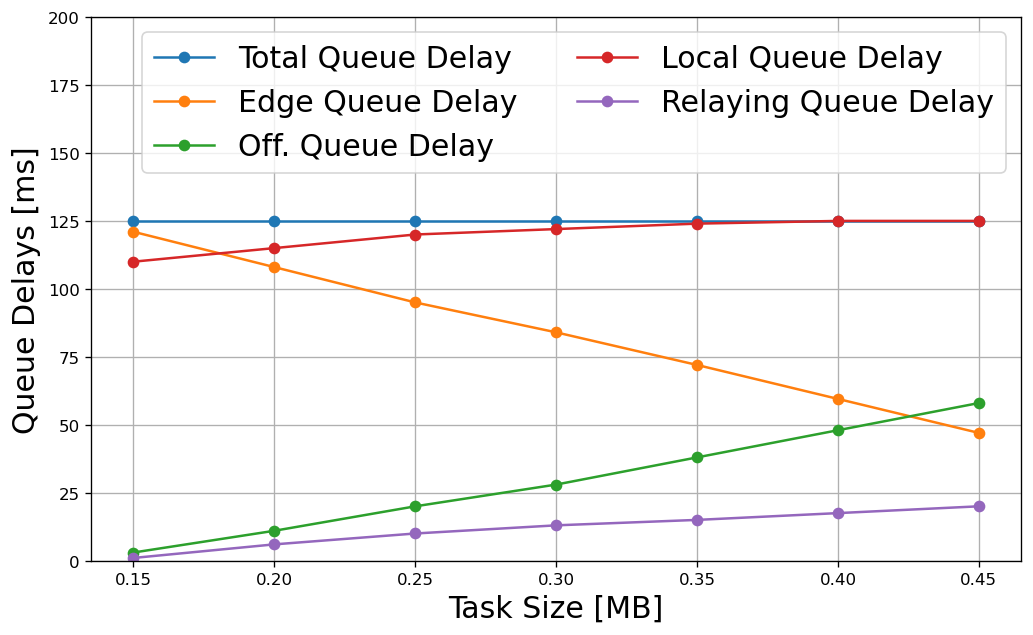}
    \caption{ $t_n$}
    \label{fig:TVsTs}
    \end{subfigure}
    \caption{ Different system parameters with task size.}
    \label{fig:task size}
\end{figure}

In this scenario, all parameter values remain consistent with those in the previous subsection, except for the number of IoTDs, $N$, which is $50$. As task size increases, the volume of tasks offloaded to the edge also increases because the local IoTDs are unable to handle the larger tasks solely with their local computing resources, as shown in Fig. \ref{fig.offVsTs}. Notably, the MAPPO-BD and PO-MAPPO-BD algorithms exhibit similar offloading ratios, as discussed in Fig. \ref{fig.offVsEpi}. However, the divergence in offloading ratios among MAPPO-BD-based algorithms and the other baselines becomes more pronounced with increasing task size.

To process these additional offloaded tasks, UAVs and HAPS must allocate more computing resources to meet task delay requirements. It is important to note that the disparity in resource allocation between MAPPO-BD and the other baselines widens as task size increases, as illustrated in Fig. \ref{fig.CompVsTs}. This indicates that MAPPO-BD manages resources more efficiently, even as the computational demand grows.

Moreover, increased offloading leads to higher transmission energy consumption. Allocating more computing resources at the edge also escalates the energy consumption of edge computing. This explains the observed trend of exponentially rising energy consumption with increasing task size, as shown in Fig. \ref{fig:EVsTs}. The MAPPO-BD algorithm, through optimized resource allocation and task processing strategies, consistently achieves lower energy consumption compared to the other baselines. The energy consumption values in this scenario represent the total energy expenditure, as described by Eq. (\ref{energy}), for all IoTDs and UAVs within our proposed network during a single time slot.

Utilizing our proposed MAPOO-BD-based algorithm, we conducted a comprehensive study of the queue delay components, including the local queue, offloading queue, edge queue, and relaying queue, in relation to task size, as illustrated in Fig. \ref{fig:TVsTs}. The delay of the local queue starts at a value of $110$ ms and increases with the task size, reaching its maximum value of $125$ ms, the maximum allowable task delay, at a task size of 0.35 MB. As the volume of offloading increases with task size, both the offloading queue delay and the relaying queue delay also rise. To respect the maximum allowable task delay \( t^{max}_n \), as offloading and relaying delays increase, the edge queue delay conversely decreases with increasing task size. This inverse relationship is justified by the fact that the exponentially increasing of the allocated computing resources at the edge compensates for the reduced edge queue delay, ensuring compliance with the delay constraints.

\begin{figure}[h]
    \centering
    \begin{subfigure}[b]{0.24\textwidth}
    \centering
    \includegraphics[width=\textwidth, height=3cm]{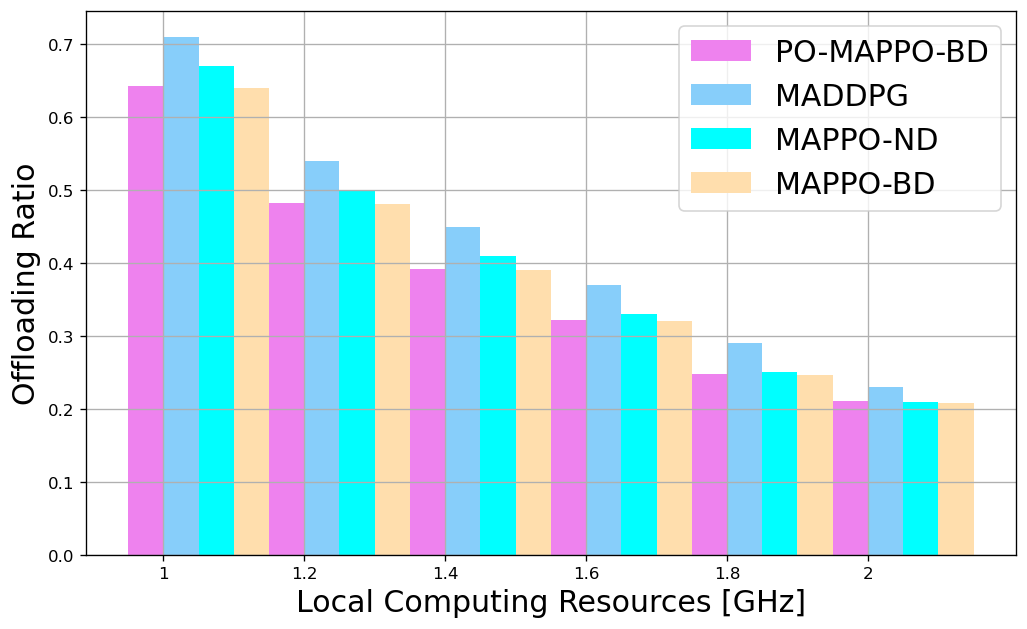}
    \caption{$\alpha_{n}^{o}$}
    \label{fig.offVsLr}
     \end{subfigure}
    \hfill
    \begin{subfigure}[b]{0.24\textwidth}
    \centering
\includegraphics[width=\textwidth,height=3cm]{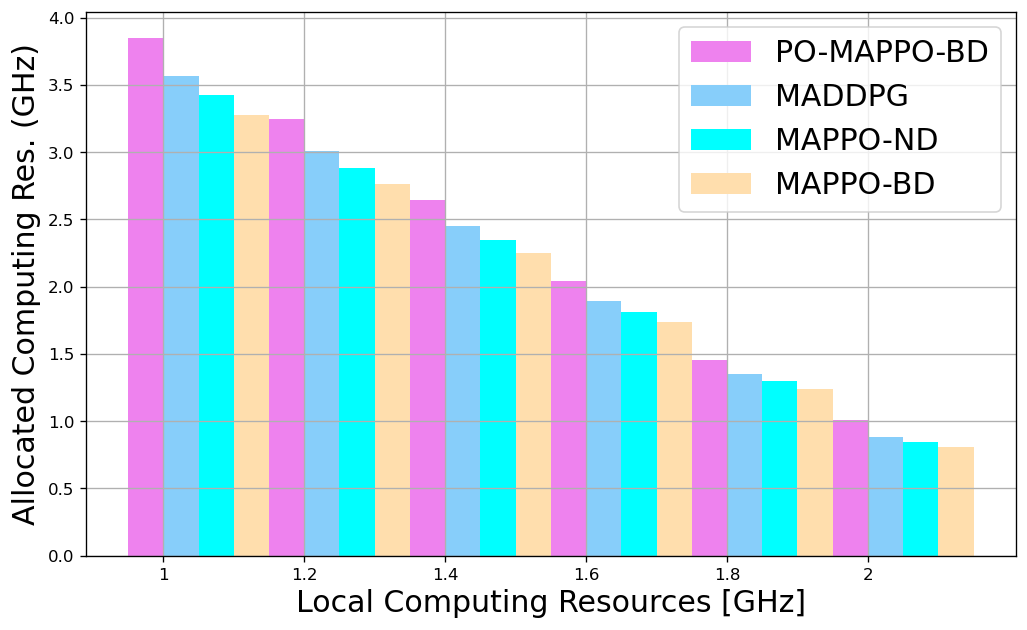}
    \caption{$f_{n}^{e}$}
    \label{fig.CompVsLr}
    \end{subfigure}
    \hfill
    \begin{subfigure}[b]{0.24\textwidth}
    \centering
\includegraphics[width=\textwidth,height=3cm]{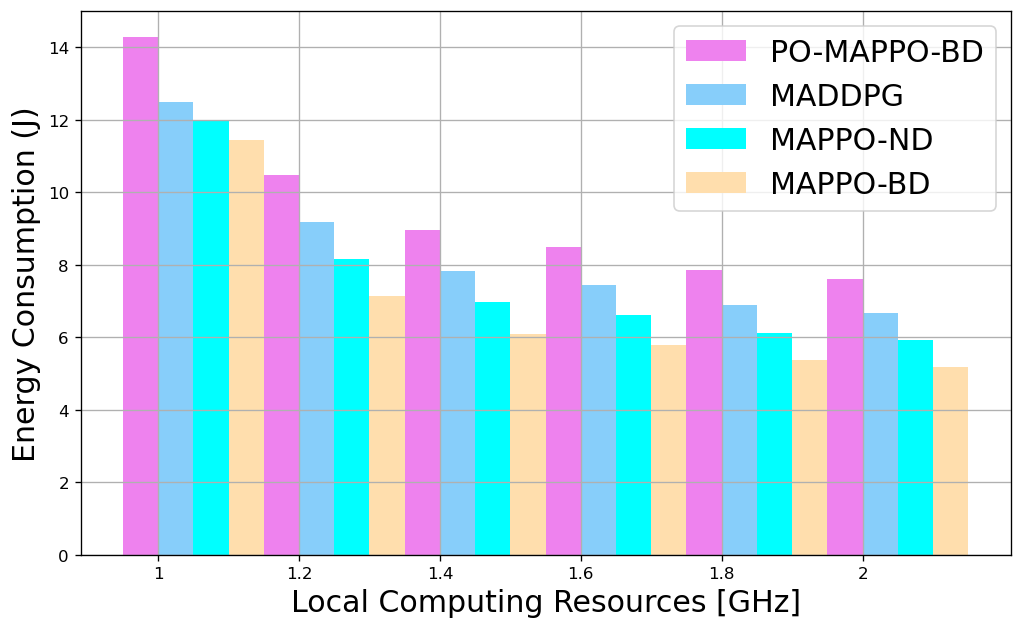}
    \caption{ $E^{all}$}
    \label{fig:EVsLr}
    \end{subfigure} 
    \hfill
    \begin{subfigure}[b]{0.24\textwidth}
    \centering
\includegraphics[width=\textwidth,height=3cm]{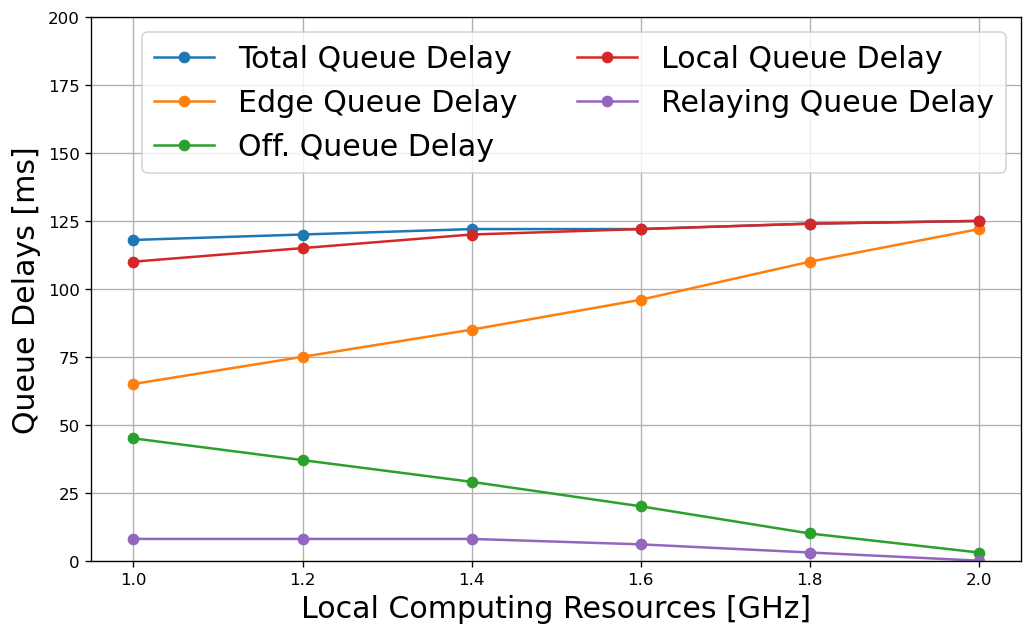}
    \caption{ $t_n$}
    \label{fig:TVsLr}
    \end{subfigure}
    \caption{ Different system parameters with local computing resources.}
    \label{fig:local resources}
\end{figure}

To evaluate our framework with more capable local IoTDs possessing higher local computing resources, another scenario was studied using the same parameter values as in the previous scenario, except for the task size, which now ranges from 0.25 to 0.35 MB. Enhancing the local computing capabilities decreases the need for offloading tasks to the edge network, as illustrated in Fig. \ref{fig.offVsLr}. Consistent with the previous analysis of task size and convergence, the MAPPO-BD and PO-MAPPO-BD algorithms exhibit similar values for the same reasons. Additionally, the allocated computing resources decrease as the offloaded load diminishes, as shown in Fig. \ref{fig.CompVsLr}. The total energy consumption within a single time slot is calculated and depicted in Fig. \ref{fig:EVsLr}, showing a reduction in energy consumption as the offloading ratio and allocated computing resources decrease with the increased local computing resources of the IoTDs. This reduction in offloading effectively conserves the energy that would otherwise be expended in transmitting tasks to the edge network.

With the same analysis of queue delays presented in Fig. \ref{fig:TVsTs}, Fig. \ref{fig:TVsLr} illustrates the queue delays when increasing local computing resources. The reduced necessity for offloading due to enhanced local computing resources results in decreased delays for both offloading and relaying queues. Additionally, to minimize energy consumption, the edge entities allocate only the minimum necessary computing resources that are sufficient to process the offloaded tasks within the maximum allowable delay \( t^{max} \).

\begin{figure}[h] 
\centering \includegraphics[width=3in, height=5.5cm]{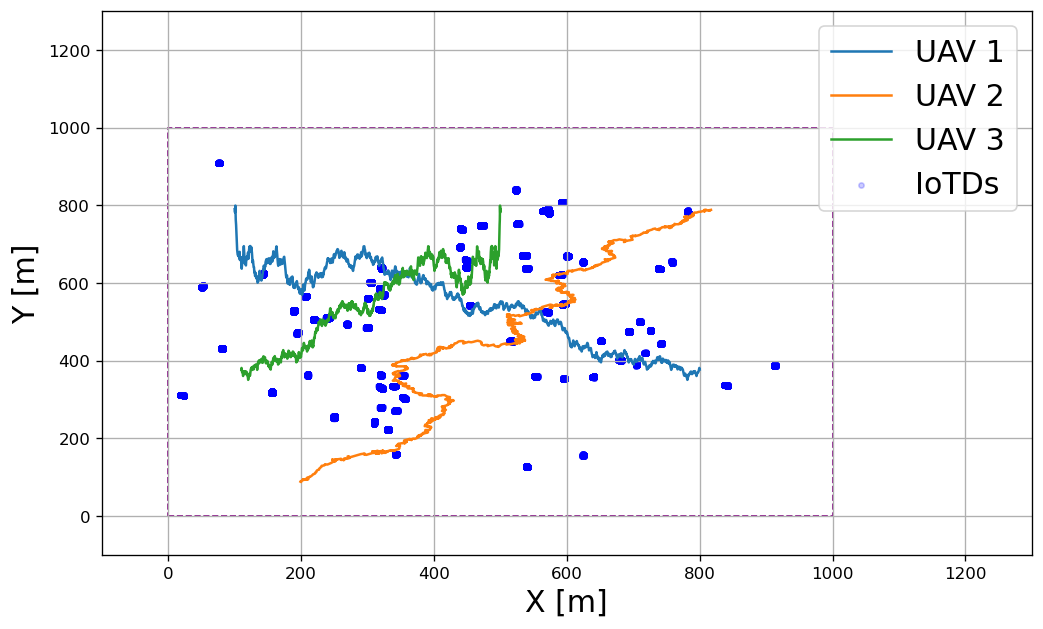} 
\caption{UAV trajectories within 10 episodes.} \
\label{fig: UAV trajec} 
\end{figure}

The initial locations of the three UAVs are [100,800], [200.,100], and [500,800], respectively. The 70 IoTDs are distributed across four hotspots, with the hotspot locations randomly chosen at the beginning of the simulation. The trajectories of the UAVs over 10 episodes are depicted in Fig. \ref{fig: UAV trajec}. These trajectories illustrate the dynamic paths the UAVs take to ensure coverage of the IoTDs. It is noteworthy that task arrivals are dynamic in each time slot, with diverse task profiles and varying delay requirements. The UAVs respond to the offloading needs of the IoTDs by prioritizing those with tight delay constraints (high priority) and subsequently allocating the appropriate computing resources to process the offloaded tasks. This dynamic path planning and resource allocation highlight the adaptive capabilities of the UAVs in managing heterogeneous and distributed IoTD environments.

\begin{figure}[h] 
\centering \includegraphics[width=3in, height=5.5cm]{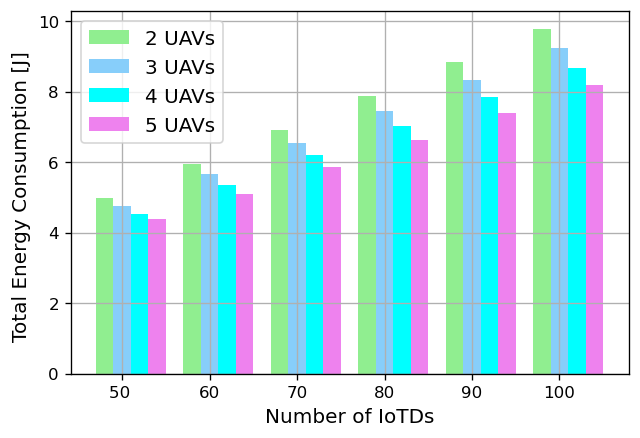} 
\caption{Total energy consumption with different $N$ and $M$.} \label{fig: NumUser} 
\end{figure}

The total energy consumption increases with the number of IoTDs for each UAV configuration, as shown in Fig. \ref{fig: NumUser}, which is expected as more IoTDs generate more tasks, requiring additional computational and transmission resources, thereby increasing energy consumption. The analysis demonstrates that increasing the number of UAVs leads to a reduction in total energy consumption, particularly as the number of IoTDs increases. This suggests that a higher number of UAVs allows for better distribution of the workload, more efficient task offloading, and improved resource utilization. This improvement can be attributed to the reduced load on the HAPS, which decreases interference among IoTDs served by the HAPS. Consequently, this enhances the data rate and decreases the time and energy required for task offloading from the IoTDs to the HAPS. The scalability of the MAPPO-BD framework is demonstrated through these scenarios, showing that the algorithm adapts well to increasing network sizes. The model maintains efficient performance even as the number of UAVs and IoTDs grows. The use of MARL inherently enhances scalability by enabling multiple agents (UAVs) to dynamically coordinate decisions and optimize resource management, ensuring the system's efficiency in large-scale environments and making it suitable for real-world network expansion.

\begin{figure}[h]
    \centering
        \begin{subfigure}{0.24\textwidth}
\includegraphics[width=\textwidth,height=3.5cm]{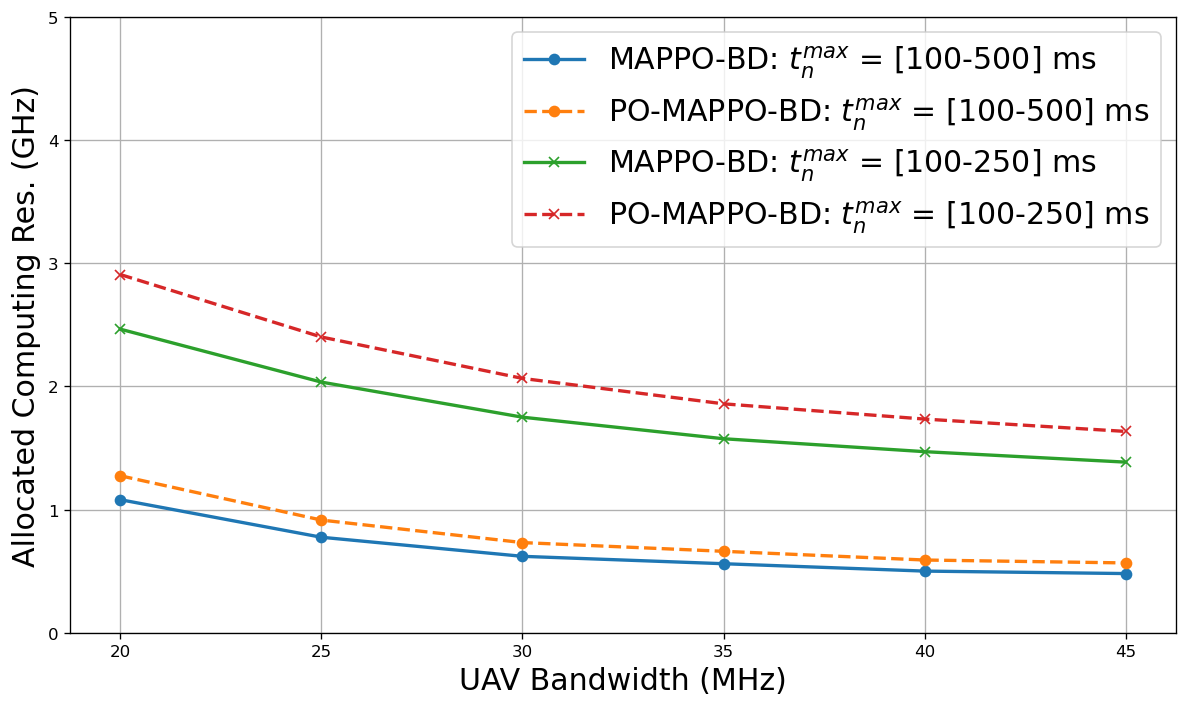} 
         \caption{$f_{n}^{e}$}
        \label{fig:CompVsBand}
        \end{subfigure}
        \begin{subfigure}{0.24\textwidth}
\includegraphics[width=\textwidth,height=3.5cm]{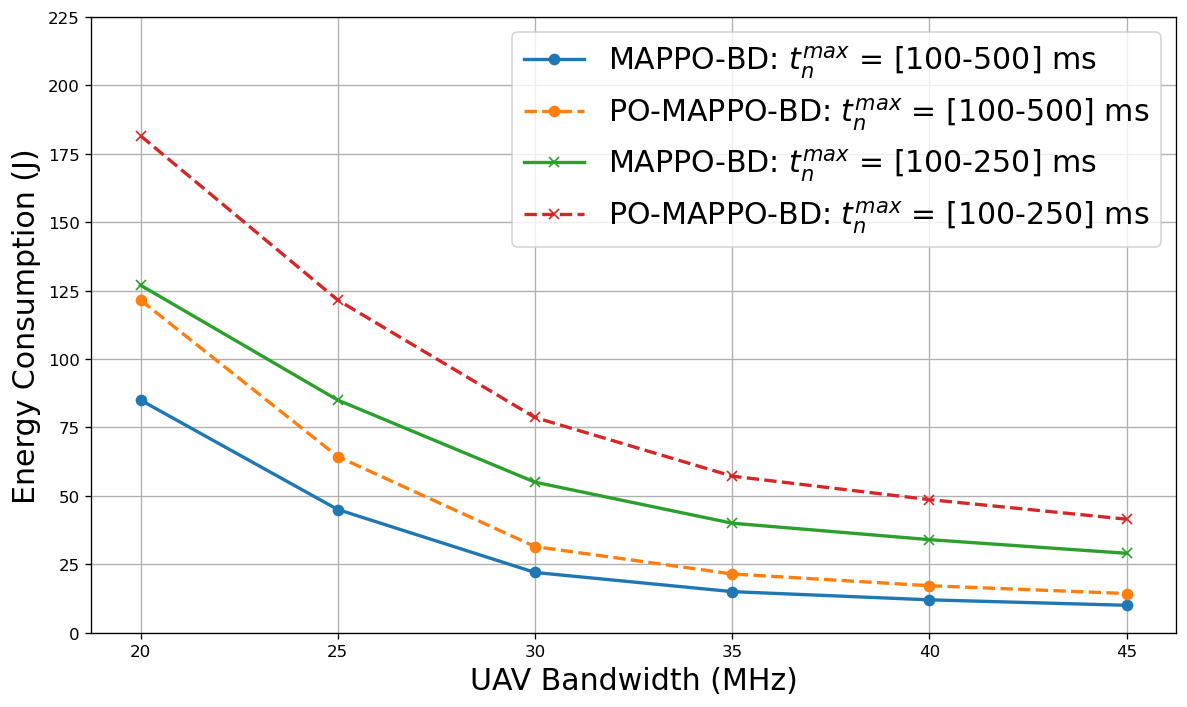} 
         \caption{$E^{all}$}
        \label{fig:EnVsBand}
        \end{subfigure}
        \caption{The edge allocated computing resources and the total energy consumption vs. frequency bandwidth and maximum allowable task delay.}
    \label{fig:en comp vs Band}
\end{figure}

Fig. \ref{fig:en comp vs Band} illustrates the impact of increasing UAV bandwidth and the maximum allowable delay on allocated computing resources and total energy consumption. Increasing UAV bandwidth significantly reduces both energy consumption (Fig. \ref{fig:EnVsBand}) and the need for allocated computing resources (Fig. \ref{fig:CompVsBand}). Higher bandwidth allocated to the UAV facilitates faster data transmission, thereby decreasing the time required for task offloading. As the offloading time decreases, the additional available time for edge computing increases, allowing for the allocation of fewer computing resources to process the offloaded tasks. This results in decreased energy consumption for both transmission and edge computing. A broader range of allowable task delays (100-500 ms) enhances the system's flexibility. This flexibility is reflected in reduced allocated computing resources and, consequently, in the lower energy consumption of edge computing, particularly for the wider delay range. The MAPPO-BD algorithm consistently outperforms the PO-MAPPO-BD algorithm by allocating fewer computing resources, resulting in significant energy savings.

While our simulations assume specific network conditions and agent behaviors, we acknowledge that these simplifications may not fully capture the variability of real-world dynamics. To assess the robustness of the proposed MAPPO-BD framework under different scenarios, we conducted a sensitivity analysis in this section, which demonstrates consistent performance across a range of conditions.

\section{Conclusion}  
In this work, we have introduced a queue-aware task offloading method and a multi-tier computing approach in MAGIN to handle the challenges posed by the dynamic task profiles and the high QoS requirements of IoTDs with different real-time applications and Metaverse. Our proposed solution minimized the energy consumption by optimizing UAV trajectories, computing resource allocation, and task offloading decisions while considering task requirements, queue latency constraints, and the maximum computing capabilities of the ABS. We leveraged dynamic HMADRL strategies and proposed a MAPPO-BD-based algorithm. It facilitated more uniform exploration and avoided probability density concentration near the boundaries, which is particularly beneficial for heterogeneous agents with varying action boundaries. Extensive simulations comparing our MAPPO-BD algorithm against baselines such as MADDPG, MAPPO-ND, and PO-MAPPO-BD confirmed the superiority of our approach. The results showed that MAPPO-BD significantly outperforms these baselines, achieving superior energy savings and more efficient resource management in MAGIN. Given that energy consumption is influenced by the task profile and the other network parameter values, our system model demonstrates that the MAPPO-BD-based approach can achieve up to 32\% greater energy savings compared to the PO-MAPPO-based approach. In future work, to improve the evaluation of the MAPPO-BD framework’s adaptability and robustness in real-world settings, we will include more dynamic network conditions, such as varying interference levels, unpredictable agent mobility, and changing environmental factors.

\bibliographystyle{IEEEtran}
\bibliography{IEEEabrv,main}
\begin{IEEEbiography}[
{\includegraphics[width=1in,height=1.25in,clip,keepaspectratio]{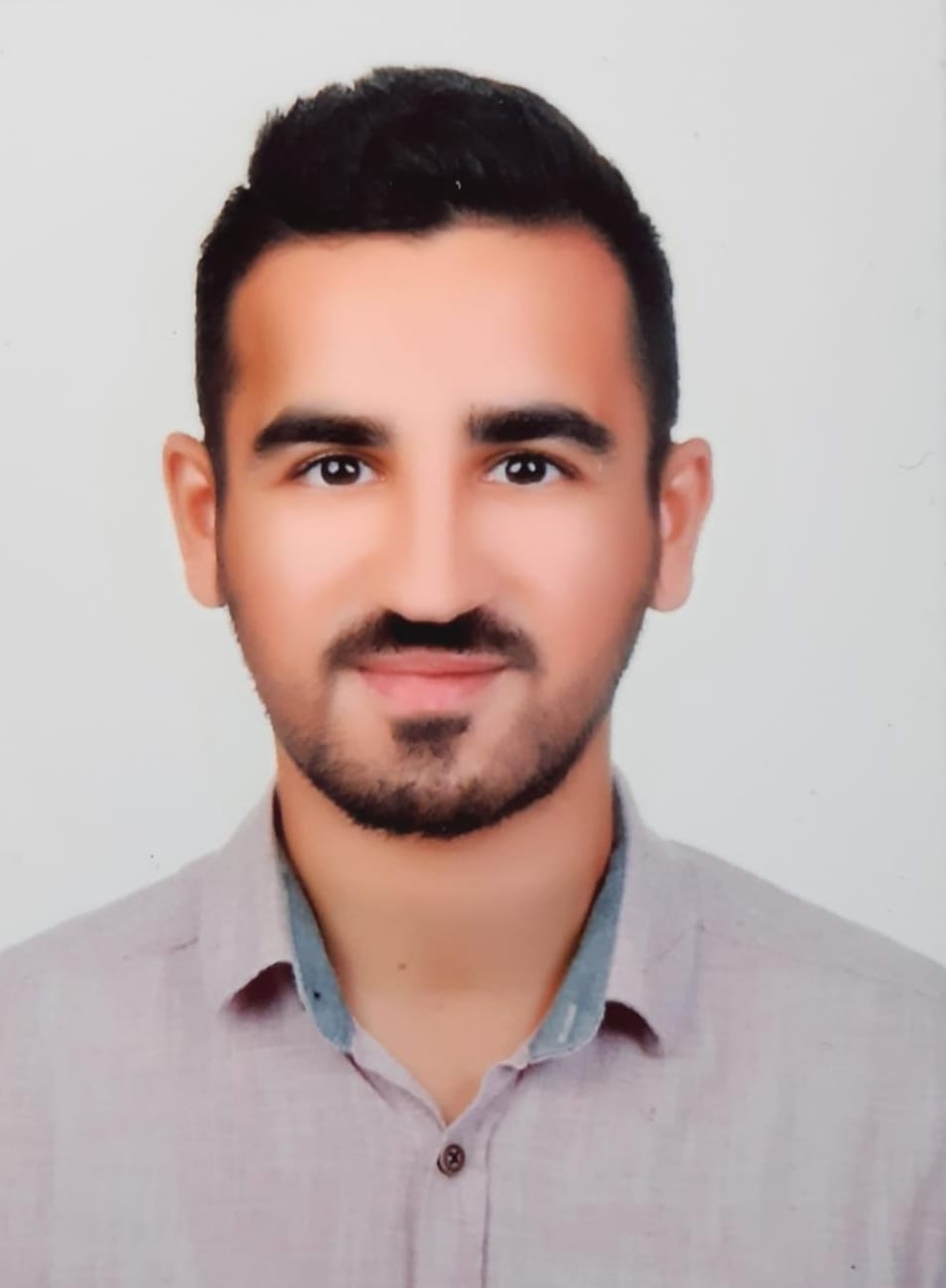}}
]
{\rmfamily Muhammet Hevesli} received his M.Sc degree in Telecommunication Engineering from Kocaeli University, Turkey, in 2019. Currently, he is a Ph.D. candidate in the Department of Information and Computing Technology at Hamad Bin Khalifa University, Qatar. Muhammet has published several scientific papers in different conferences and transactions. He is a member of IEEE and also served as a member of the TPC at various IEEE conferences like WCNC, ISNCC, and IWCMC. His research interests include unmanned aerial vehicles, internet of things, wireless communications, deep reinforcement learning, and edge computing. 
\end{IEEEbiography}

\begin{IEEEbiography}[
{\includegraphics[width=1in,height=1.25in,clip,keepaspectratio]{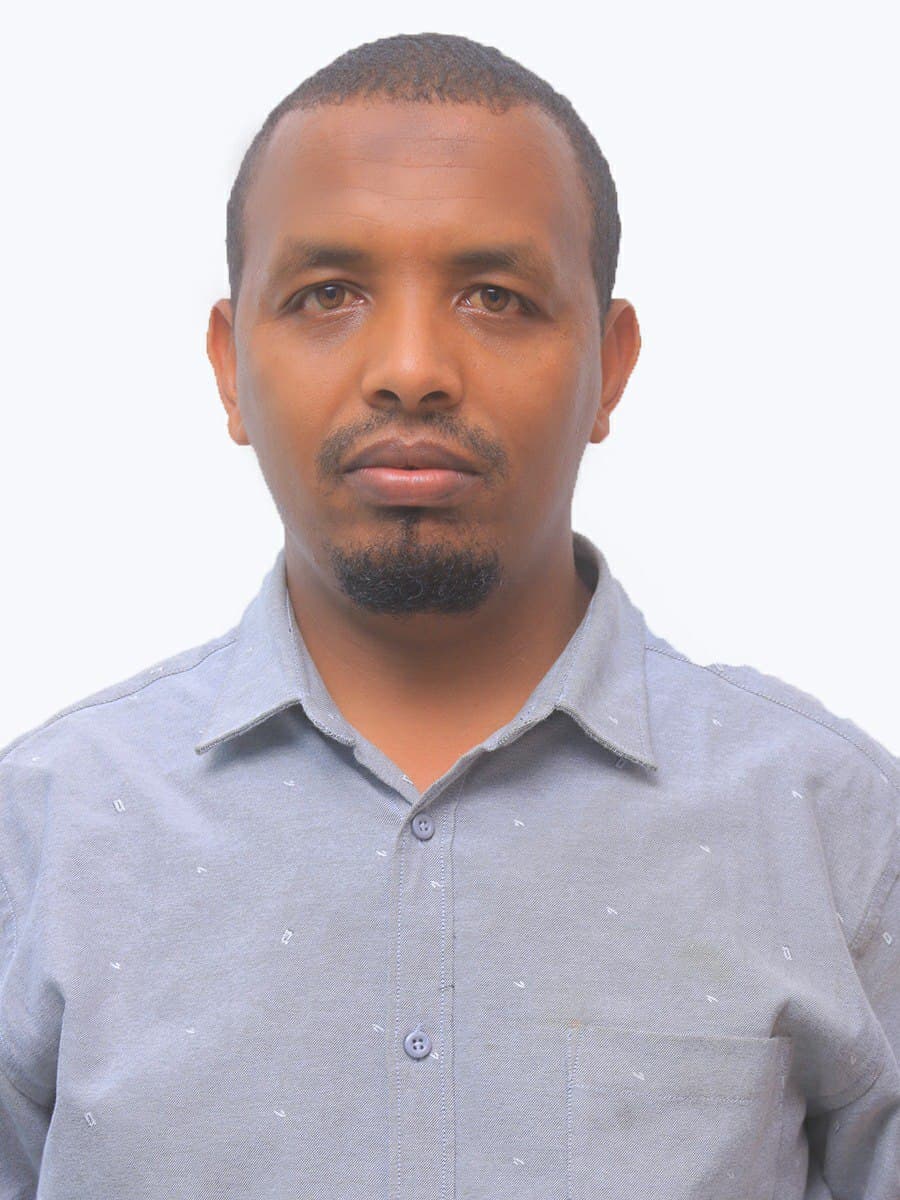}}
]
{\rmfamily Abegaz Mohammed Seid} received his B.Sc. and M.Sc. degrees in Computer Science from Ambo University and Addis Ababa University, Ethiopia, in 2010 and 2015, respectively. He received a Ph.D. degree in Computer Science and Technology from the University of Electronic Science and Technology of China (UESTC) in 2021. He is currently a post-doctoral fellow with the College of Science and Engineering at Hamad Bin Khalifa University, Doha, Qatar. Dr. Abegaz has published more than thirty scientific conferences and journal papers. His research interests include wireless networks, mobile edge computing, blockchain, machine learning, vehicular networks, IoT, machine learning, UAV networks, IoT, and 5G/6G wireless networks. He is a member of IEEE.
\end{IEEEbiography}

\begin{IEEEbiography}[
{\includegraphics[width=1in,height=1.25in,clip,keepaspectratio]{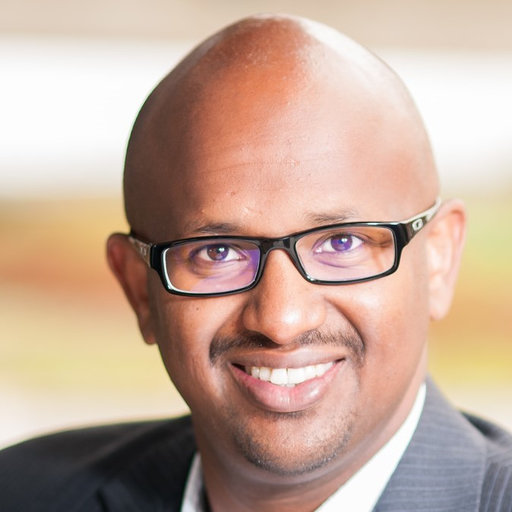}}
]
{\rmfamily Aiman Erbad (aerbad@ieee.org)} is a Professor and VP Research and Graduate Studies at Qatar University. Dr. Erbad obtained a Ph.D. in Computer Science from the University of British Columbia (Canada) in 2012. He received the Platinum award from H.H. The Emir Sheikh Tamim bin Hamad Al Thani at the Education Excellence Day 2013 (Ph.D. category). He is an editor for KSII Transactions on Internet and Information Systems, an editor for the International Journal of Sensor Networks (IJSNet), and a guest editor for IEEE Network. He also served as a member of the Technical Program Committee (TPC) at various IEEE and ACM international conferences. His research interests span cloud computing, edge intelligence, Internet of Things (IoT), private and secure networks, and multimedia systems. He is a senior member of IEEE and ACM. 
\end{IEEEbiography}

\begin{IEEEbiography}[{\includegraphics[width=1in,height=1.25in,clip,keepaspectratio]{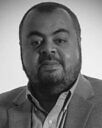}}
]
{\rmfamily Mohamed Abdallah} is currently a Founding Faculty Member with the rank of Professor at the College of Science and Engineering, Hamad Bin Khalifa University, Doha. He received his M.Sc. and Ph.D. degrees from the University of Maryland at College Park, MD, USA, in 2001 and 2006, respectively. From 2006 to 2016, he held academic and research positions with Cairo University and Texas A \& M University in Qatar, Doha, Qatar. His current research interests include wireless networks, wireless security, smart grids, optical wireless communication, and blockchain applications for emerging networks. His professional activities include being an Associate Editor of the IEEE Transactions on Communications and the IEEE Open Access Journal of Communications and a technical program committee member of several major IEEE conferences. \end{IEEEbiography}
\end{document}